\documentclass[11pt]{article}

\pdfoutput=1
\usepackage{xargs}

\usepackage{lipsum}
\usepackage{amsmath}% http://ctan.org/pkg/amsmath
\usepackage{amssymb}
\usepackage{graphicx, subcaption}
\usepackage{color}
\usepackage{chngpage}

\usepackage[backend=bibtex, style=ieee]{biblatex}
\bibliography{main}

\usepackage{tablefootnote}

\usepackage{csquotes}
\usepackage[noend]{algpseudocode}
\usepackage{algorithmicx}
\usepackage{algorithm}
\usepackage{authblk}

\usepackage{ntheorem}
\theoremseparator{:}
\newtheorem{thm}{Theorem}

\newtheorem{hyp}{Hypothesis}
\newtheorem{proof}{Proof}

\algdef{SE}[DOWHILE]{Do}{doWhile}{\algorithmicdo}[1]{\algorithmicwhile\ #1}%

\newcommand{\enotesoff}{\long\gdef\enote##1{}}
\newcommand{\enoteson}{\long\gdef\enote##1{\par\noindent\fbox{\parbox {\textwidth}{{\large DRAFT.} \small\scshape ##1}}\\[0.3ex]}}
\enotesoff
\enoteson

\newcommand{\potamnoteson}{\long\gdef\potamnote##1{\par\noindent\fbox{\parbox {\textwidth}{{\large potam comment -- draft.} \small\scshape ##1}}\\[0.3ex]}}
\potamnoteson

\newcommand{\thyenoteson}{\long\gdef\thynote##1{\par\noindent\fbox{\parbox {\textwidth}{{\large thymios comment -- draft.} \small\scshape ##1}}\\[0.3ex]}}
\thyenoteson

\newcommand{\geoenoteson}{\long\gdef\geonote##1{\par\noindent\fbox{\parbox {\textwidth}{{\large george comment -- draft.} \small\scshape ##1}}\\[0.3ex]}}
\geoenoteson

\newcommand{\specificthanks}[1]{\@fnsymbol{#1}}
\usepackage{hyperref}
\hypersetup{
    colorlinks=true,
    linkcolor=black,
    filecolor=black,      
    urlcolor=blue,
}
 
\usepackage{amsmath}
\author[1]{Georgios Paraskevopoulos\footnote{Georgios Paraskevopoulos and Efthymios Tzinis
contributed equally}\footnote{corresponding author
\href{mailto:geopar@central.ntua.gr}{geopar@central.ntua.gr} }}
\author[1]{Efthymios Tzinis\textsuperscript{*}}
\author[2]{Emmanuel-Vasileios Vlatakis-Gkaragkounis}
\author[1]{Alexandros Potamianos}

\affil[1]{National Technical University of Athens}
\affil[2]{Columbia University in the City of New York}

\begin{document}
\title{
Pattern Search Multidimensional Scaling
}

% \author{\name Georgios Paraskevopoulos\thanks{G. Paraskevopoulos and E. Tzinis contributed equally} \email geopar@central.ntua.gr \\
%        \addr Department of Electrical and Computer Engineering \\
%        National Technical University of Athens \\
%        \AND
%        \name Efthymios Tzinis\footnotemark[1] \email etzinis@gmail.com \\
%        \addr Department of Electrical and Computer Engineering \\
%        National Technical University of Athens \\
%        \AND 
%        \name Emmanouil Vasileios Vlatakis Gkaragkounis \email emvlatakis@cs.columbia.edu \\
%        \addr Department of Computer Science \\
%        Columbia University
%        \AND
%        \name Alexandros Potamianos \email potam@central.ntua.gr \\
%        \addr Department of Electrical and Computer Engineering \\
%        National Technical University of Athens
%        }

\maketitle

%%%%%%%%%%%%%%%%%%%%%%%%%%%%%%%%%%%%%%%%%%%%%%%%%%%%%%%%%%%%%%%%%%%%%%%%%%%%%%%%
\begin{abstract}

We present a novel view of nonlinear manifold learning using derivative-free optimization techniques.
Specifically, we propose an extension of the classical multi-dimensional scaling (MDS) method, where instead of performing gradient descent, we sample and evaluate possible ``moves'' in a sphere of fixed radius for each point in the embedded space. A fixed-point convergence guarantee can be shown by formulating the proposed algorithm as an instance of General Pattern Search (GPS) framework. 
Evaluation on both clean and noisy synthetic datasets shows that pattern search MDS can accurately infer the intrinsic geometry of manifolds embedded in high-dimensional spaces. 
Additionally, experiments on real data, even under noisy conditions, demonstrate that the proposed pattern search MDS yields state-of-the-art results. 
\end{abstract}

%%%%%%%%%%%%%%%%%%%%%%%%%%%%%%%%%%%%%%%%%%%%%%%%%%%%%%%%%%%%%%%%%%%%%%%%%%%%%%%%
\section{INTRODUCTION
\label{section:INTRODUCTION}
}
In the past decades, we have been witnessing a steady increase in the size of datasets generated and processed
by computational systems.  Such voluminous data comes from various sources, such as business sales records, the collected results of scientific experiments or real-time sensors used in the Internet of Things (IoT). The most popular way to represent such data is via a set of data points lying in a vector space. The construction of the vector space is often performed using a distance or similarity matrix that can be constructed manually using perceptual ratings or, more commonly, computed automatically using a set of features. 
%Fortunately, in many of these applications, even if the data-vectors are high-dimensional, one could consider them to be lying on or close to an embedded low-dimensional, not necessarily linear, manifold, which is as the manifold hypothesis \cite{fefferman2016testing}.  %in a high-dimensional ambient space. 
In many of these applications high-dimensional data representations are assumed to lie in the vicinity of a low-dimensional, possibly non-linear manifold, embedded in the high-dimensional space. This is known as the manifold hypothesis \cite{fefferman2016testing}.
%\thynote{I do not like this sentence at all. I would prefer sth like: Even if the representations of the data are high-dimensional vectors, the underlying distribution of the data could be also be approximated by a nonlinear manifold embedded in a lower-dimensional space.}
Intuitively human cognition also performs similar mappings when performing everyday tasks, i.e.,   
%\potamnote{you attempt to discuss sparsity next in a probably naive way ... needs to be reworked}
high-dimensional sensory input get embedded into low dimensional cognitive subspaces \cite{kanerva1988sparse}; \cite{pothos2011quantum}; \cite{pothos2013can} for rapid and robust decision making, since only a small number of features are salient for each task. Given this assumption
%the 
\textit{manifold learning} 
%task 
aims to discover such hidden low-dimensional structure and to output a representation with much fewer ``intrinsic variables''.
%\potamnote{george we need a couple of references here}

%\potamnote{why non-metric ... the proposed algorithm is general ... what you mean to say here is probably that we propose a new algorithm and investigate its performance in non metric ...}
In this paper, we study the problem of manifold learning in non-metric topological spaces. The input to this problem is a matrix of (similarities or) dissimilarities\footnote{It should be mentioned that in many real-world tasks the used dissimilarity measures may correspond in pseudo- or semimetric distance functions that violate the triangular inequality.} of the dataset objects. ``Objects'' can be colors, faces, map coordinates, political persuasion scores, or any kind of real-world or synthetic stimuli. 
%\potamnote{should this be ordinal?}
For each input dataset object, the output is a low-dimensional vector such that the pairwise Euclidean  distances of the output vectors resemble the original dissimilarities. This problem is known as non-metric multidimensional scaling (MDS) or non-linear dimensionality reduction (NLDR) task. An abundance of embedding methods have been developed for dealing with this task as detailed in Section~\ref{section:RELATED WORK}. 

The majority of these algorithms reduce this problem to the optimization of a deterministic loss function $f$. Given this minimization objective, they usually employ gradient-based methods to find a global or a local optimum. In many situations, however, the loss function is non-differentiable or estimating its gradient may be computational expensive. Additionally, gradient-based algorithms usually yield a slow convergence; multiple iterations are needed in order to minimize the loss function. 

Inspired by the recent progress in derivative-free optimization tools, we propose an iterative algorithm which treats the non-metric MDS task as a derivative-free optimization problem. The main contributions of the paper are as follows: 1) Using the General Pattern Search (GPS) formulation we are able to provide theoretical convergence guarantees for the proposed non-metric MDS algorithm. 2) A set of heuristics are proposed that significantly improve the performance of the proposed algorithm in terms of computational efficiency, convergence rate and solution accuracy. 3) The proposed algorithm is evaluated on a variety of tasks including manifold unfolding, word embeddings and optical digit recognition, showing consistent performance and good convergence properties. We also compare performance with state-of-the-art MDS algorithms for the aforementioned tasks for clean and noisy datasets. An optimized implementation of pattern search MDS and the experimental code is made available as open source to the research community\footnote{Open source code available: \href{https://github.com/georgepar/pattern-search-mds}{https://github.com/georgepar/pattern-search-mds}}. 

The remainder of the paper is organized as follows:
We begin with an overview of the related work in Section~\ref{section:RELATED WORK}. Then in Section~\ref{section:PRELIMINARIES},
we review several optimization problems that are related to the manifold learning task and we present the 
GPS framework. % where the proposed MDS-like algorithm falls into. 
Then in Section \ref{section:OUR ALGORITHM}, we present in detail the proposed derivative-free algorithm, a sketch of the reduction of the algorithm to the GPS formulation and the associated proof of fixed-point convergence guarantees.
Finally in Section~\ref{section:EXPERIMENTS}, the proposed algorithm is compared and contrasted with other dimensionality reduction methods in a variety of tasks with or without the presence of noise. We conclude and present future directions for research in Sections~\ref{section:CONCLUSIONS} and~\ref{section:FUTURE WORK}, respectively.

%The problem of multidimensional scaling, broadly stated, is to find $n$
%points whose interpoint distances match in some sense the experimental
%dissimilarities of $n$ objects.However, in many data analysis tasks, one is often confronted with very high dimensional data.
%There is a strong intuition that the data may have a lower dimensional intrinsic representation.

%Recently, we have witnessed a colossal increment in the sizes of data sets generated and processed
%by computing systems. Yet there are many sources which are all in broad agreement that the size of the digital universe will double every two years at least, a 50-fold growth from 2010 to 2020 \cite{?}. As the volume of the data increases, memory and processing
%requirements need to correspondingly increase at a similar fast pace, and this is frequently restrictively 
%costly. Consequently, there has been considerable interest in the task of effective modeling of high-dimensional observed data and information; such models must encorporate the structure of the information content in a concise manner.

%While trying to mitigate such issues, a new model as far as \textit{manifolds} for finding relevant features and representing the data by a few parameters
%is gaining interest by machine learning and signal processing communities. Especially, while early AI's problem focused on the grand goals of building machines that mimicked the human brain, and some underlying cognitive truth exists, a construction of a low-dimensional manifold is working toward this goal.
%

\section{RELATED WORK \label{section:RELATED WORK}}
% \potamnote{AP maybe replace Euclidean with normed space or Hilbert space? Is our work limited to real numbers with the L2 norm?}
% \thynote{Some sentences presented in this paragraph are a bit abstract for the \textbf{RELATED WORK} section. Especially, the following sentence should be in the introduction I think. In related work I would like to see a more detailed description of the related algorithms and not abstract topological spaces}
% Loosely speaking, a manifold is a topological space that locally resembles a Euclidean space. When data lies on or close to a globally linear subspace, the problem of manifold learning can be reduced to the classical linear dimensionality reduction task. The most popular algorithms for this problem are Principle Components Analysis (PCA) \cite{Pearson} and Multidimensional Scaling (MDS) \cite{ClassicalMDS}. PCA minimizes the dimension of the final representation, preserving the covariance of data. MDS aims to preserve the distances between data points.

Loosely speaking, a manifold is a topological space that locally resembles a Euclidean space. The purpose of Multidimensional Scaling (MDS) is to infer data representations on a low-dimensional manifold while simultaneously preserving the distances of the high-dimensional data points. When data lies on or close to a linear subspace, low-dimensional representations of data can be obtained using linear dimensionality reduction techniques like Principle Components Analysis (PCA) \cite{Pearson} and classical MDS.

% \thynote{You never know if the data are lying on a manifold you only assume that you can approximate their distribution by learning a manifold.}

% \thynote{This is totally wrong, there is no such task. PCA is a linear dimensionality reduction technique and MDS is a form of non linear dimensionality reduction... }

% \thynote{These sentences are just not connected and should not represented in this way}

In real data applications, such a linearity assumption may be too strong and can lead to meaningless results. Thus a significant effort has been made by the machine learning community to apply manifold learning in non-linear domains.
Representative manifold learning algorithms include Isometric Feature Mapping (ISOMAP) 
\cite{tenenbaum_global_2000,bernstein_graph_2000,zha_isometric_2003,DonohoG05,pless_image_2003}, 
Landmark ISOMAP \cite{silva2004-landmark,silva_global_2003},Locally Linear Embedding (LLE)
\cite{BelkinN03,Cayton,saul_think_2003,sha_analysis_2005, belkin_laplacian_2001},
Modified LLE
\cite{zhang2007mlle}
Hessian LLE
\cite{zhang_principal_2004,donoho_hessian_2003},
Semidefinite Embedding \cite{weinberger2006unsupervised},
\cite{weinberger2005nonlinear},
\cite{vandenberghe_semidefinite_1996},
\cite{bertsekas_nonlinear_1999},
Laplacian Eigenmaps (LE) \cite{BelkinN01,BelkinN03,BelkinN06}, Local Tangent Space Alignment (LTSA) \cite{LTSA},etc. 
%\potamnote{ this is too cryptic and short - needs to be expanded and the intuition behind the last comment needs to be verified via a citation}
ISOMAP uses a geodesic distance to measure the geometric
information within a manifold. LLE assumes that a 
manifold can be approximated in a Euclidean
space and the reconstruction coefficients of neighbors
can be preserved in the low-dimensional space. LE uses
 an undirected weighted graph to preserve local neighbor
relationships. Hessian LLE obtains low-dimensional representations
through applying eigenanalysis on a  Hessian coefficient matrix.
LTSA utilizes local tangent information to represent the
manifold geometry and extends this to global coordinates.
Finally, SDE attempts to maximize the distance between points that don't belong in a local neighborhood. %If the manifold looks like a curved version of a parameter space, then hopefully this procedure will unroll it properly. %Unfortunately, the easieast way to describe this process is by defining a semidefinite program implying huge computational restrictions. 
Also, a common nonlinear method for dimensionality reduction is the kernel extension of PCA \cite{scholkopf1998nonlinear}.

A wide class of derivative-free algorithms for nonlinear optimization has been studied and analyzed in \cite{Rios2013} and \cite{avriel2003nonlinear}. GPS methods consist a subset of the aforementioned algorithms which do not require the explicit computation of the gradient in each iteration-step. Some GPS algorithms are: the original Hooke and Jeeves pattern search algorithm \cite{Hooke:1961:DSS:321062.321069}, the evolutionary operation by utilizing factorial design \cite{box1957evolutionary} and the multi-directional search algorithm \cite{torczon1989multidirectional}, \cite{doi:10.1137/0801027}. In \cite{torczon1997convergence}, a unified theoretical formulation of GPS algorithms under a common notation model has been presented as well as an extensive analysis of their global convergence properties. Local convergence properties have been studied later by \cite{dolan2003local}. Notably, the theoretical framework as well as the convergence properties of GPS methods have been extended in cases with linear constrains \cite{doi:10.1137/S1052623497331373}, boundary constrains \cite{doi:10.1137/S1052623496300507} and general Lagrangian formulation \cite{doi:10.1137/0728030}.

%\potamnote{Maybe a small paragraph that compares the aforementioned techniques with ours should be added. We should add the advantages and disadvantages of derivative free vs gradient descent models in terms of convergence, performance on manifolds and performance w. noise (if available).}

% \geonote{@thymios @manolis We should also talk a bit about the techniques relying on eigenvalue calculation and comment on numerical stability. This is an important feature of our method
% \thynote{Totally agree but we have not underlined this feature of our algorithm in the next sections
% \geonote{Check now in \ref{ss:geometry} 3d-clusters }}
% \geonote{This is shown when the Hessian/Modified LLE don't run for certain tasks. Also the 3d clusters experiment is designed to show that (bc of the sparse distance matrices). I just need to add a couple of sentences there.}
% }

%In this paper we analyze a renewed line of research where traditional algorithms are combined with derivative-free optimization toolbox in order to increase their computational efficiency.

%\section{RELATED WORK}

\section{PRELIMINARIES \label{section:PRELIMINARIES}  
}
%\subsection{Dimensionality Reduction}
%I think this should be a part of intro and not preliminaries. Very good paper for review in DR (2009) Dimensionality Reduction Comparative Review \cite{Maaten08dimensionalityreduction} -thymios

% \geonote{General comment: I noticed in preliminaries we use the word "we" a lot which is inaccurate. It should be replaced with smth like the authors, or X et al.
% \thynote{UPDATE: Fixed everything -- pls recheck}}

% \geonote{Capitalize ref statements. E.g.

% in Eq. \ref{eq gram matrix}: in Eq. \ref{eq gram matrix}

% in algorithm \ref{alg: GPS algorithm}: in Alg. \ref{alg: GPS algorithm}
% \thynote{FIXED -- Also Changed the following: \\ 
% Hypothesis $\rightarrow$ Hyp. \\
% Theorem $\rightarrow$ Thm.}
% }

\subsection{Notation}
We denote real, integer and natural numbers as $\mathbb{R}$, $\mathbb{Z}$, $\mathbb{N}$, respectively. Scalars are represented by no-boldface letters, vectors appear in boldface lowercase letters and matrices are indicated by boldface uppercase letters. All vectors are assumed to be column vectors unless they are explicitly defined as row vectors. For a vector $\mathbf{z} \in \mathbb{R}^{n}$, $\|\mathbf{z}\|_1 = \sum_{i=1}^{n}|z_i|$ is its ${\ell}_1$ norm and $\|\mathbf{z}\|_2 = \sqrt[]{\sum_{i=1}^{n}z_i^2}$ is its ${\ell}_2$ norm, where $z_i$ is the ith element of $\mathbf{z}$. By $\mathbf{A} \in \mathbb{R}^{n \times m}$ we denote a real-valued matrix with n rows and m columns. Additionally, the $j$th column of the matrix $\mathbf{A}$ and its entry at $i$th row and $j$th column are referenced as $\mathbf{a}_j$ and $a_{ij}$, respectively. The trace of the matrix $\mathbf{A}$ appears as $tr(\mathbf{A})$ and its Frobenius norm as $||\mathbf{A}||_F = \sqrt[]{\sum_{i=1}^{n}\sum_{j=1}^{m}a_{ij}^2}$.  The square identity  matrix with $n$ rows is denoted as $\mathbf{I}_n \in \mathbb{R}^{n \times n}$. For the matrices $\mathbf{A} \in \mathbb{R}^{n \times m}$ and $\mathbf{B} \in \mathbb{R}^{n \times m}$ we indicate their Hadamard product as $\mathbf{A} \odot \mathbf{B}$. The $n$-ary Cartesian product over $n$ sets $S_1,...,S_n$ is denoted by $\{(s_1,...,s_n): s_i \in S_i. \enskip 1 \leq i \leq n\}$ Finally, $\mathbf{X}^{(k)}$ refers to the estimate of a variable $\mathbf{X}$ at the $k$th iteration of an algorithm.  

% \potamnote{Should you define the rest of your notation here also, e.g., Frobenius norm, and Hadamard product? $\mathbf{X}^{(n)}$ refers to the estimate of a variable $\mathbf{X}$ at the $n$th iteration of an algorithm.
% \thynote{DONE -- I have also included all the other notation that we use}}

\subsection{Classical MDS}
Classical MDS was first introduced by \cite{ClassicalMDS} and can be formalized as follows. Given the matrix $\mathbf{\Delta}$ consisting of pairwise distances or dissimilarities $\{\delta_{ij}\}_{1 \leq i , j \leq N}$ between $N$ points in a high dimensional space, the solution to Classical MDS is given by a set of points $\{\mathbf{x}_i\}_{i=1}^N$ which lie on the manifold $\mathcal{M} \in \mathbb{R}^{L}$ and their pairwise distances are able to preserve the given dissimilarities $\{\delta_{ij}\}_{1 \leq i , j \leq N}$ as faithfully as possible. Each point $\mathbf{x}_i \in \mathbb{R}^{L}, \enskip 1 \leq i \leq N$ corresponds to a column of the matrix $\mathbf{X}^T \in \mathbb{R}^{L \times N}$.
The embedding dimension $L$ is selected as small as possible in order to obtain the maximum dimensionality reduction but also to be able to approximate the given dissimilarities $\delta_{ij}$ by the Euclidean distances $d_{ij}(\mathbf{X})=||\mathbf{x}_i - \mathbf{x}_j||_2 = \sqrt{\sum_{k=1}^L (x_{ik}-x_{jk})^2} $ in the embedded space $\mathbb{R}^{L}$. 
%Let $\mathbf{D}$ be the matrix of squared Euclidean distances between all $N$ points of the embedded space, such as $ d_{ij}^2 = \sum_{k=1}^L (x_{ik}-x_{jk})^2$. 

The proposed algorithm uses a centering matrix $\mathbf{H}=\mathbf{I}_N - \frac{1}{N} \mathbf{1}_N^T \mathbf{1}_N  $ in order to subtract the mean of the columns and the rows for each element. Where $\mathbf{1}_N = [1,1,...,1]$ a vector of ones in $\mathbb{R}^N$ space. By applying the double centering to the Hadamard product of the given dissimilarities, the Gram matrix $\mathbf{B}$ is constructed as follows:

\begin{equation}
\label{eq gram matrix}
\mathbf{B} = - \frac{1}{2}\mathbf{H}^T (\mathbf{\Delta} \odot \mathbf{\Delta}) \mathbf{H} 
\end{equation}
It can be shown (Ch. 12 \cite{borg_groenen_2005}) that classical MDS minimizes the Strain algebraic criterion in Eq. \ref{eq Strain} below:
\begin{equation}
\label{eq Strain}
||\mathbf{X}\mathbf{X}^T-\mathbf{B}||_F^2
\end{equation}
The eigendecomposition of the symmetric matrix $\mathbf{B}$ gives us $\mathbf{B}= \mathbf{V} \mathbf{\Lambda}\mathbf{V}^T$ and thus the new set of points consisting the embedding in $\mathbb{R}^L$ are given by the first $L$ positive eigenvalues of $\mathbf{\Lambda}$, namely $\mathbf{X}=\mathbf{V}_L$. This solution provides the same result as Principal Component Analysis (PCA) applied on the vector in the high dimensional space \cite{doi:10.1093/biomet/53.3-4.325}.
%This method is called principal coordinate analysis. 
Classic MDS was originally proposed for dissimilarity matrices $\mathbf{\Delta}$ which can be embedded with good approximation accuracy in a low-dimensional Euclidean space. However, matrices
%the target matrix $\mathbf{D}$ defines a Euclidean space ... (comment by AP: a matrix does not define a Euclidean space, a Euclidean space like any other space consists of data + metric ... the Euclidean space can be deduced from $D$ w/o error ... other spaces (other explanations of the data is also possible)
which correspond to embeddings in Euclidean sub-spaces \cite{10.1007/978-3-642-46900-8_44}, Poincare disks \cite{Poincare:MDS} and constant-curvature Riemannian spaces \cite{lindman1978constant} have also been studied. 

\subsection{Metric MDS}
Metric MDS describes a superset of optimization problems containing classical MDS. Shepard has introduced heuristic methods to enable transformations of the given dissimilarities $\delta_{ij}$ \cite{Shepard1962}, \cite{Shepard1962b} but did not provide any loss function in order to model them \cite{groenen2014past}. Kruskal in \cite{Kruskal:a} and \cite{Kruskal:b} formalized the metric MDS as a least squares optimization problem of minimizing the non-convex Stress-1 function defined in Eq. \ref{eq stress1} shown next:

\begin{equation}
\label{eq stress1}
\sigma_1(\mathbf{X},\hat{\mathbf{D}}) = 
\sqrt[]{\frac{\sum_{i=1}^{N} \sum_{j=1}^{N} (\hat{d_{ij}}-d_{ij}(\mathbf{X})) }{\sum_{i=1}^{N} \sum_{j=1}^{N} d_{ij}^2( \mathbf{X}) }}
\end{equation}

\noindent
where matrix $\hat{\mathbf{D}}$ with elements $\hat{d_{ij}}$ represents all the pairs of the transformed dissimilarities $\delta_{ij}$ that are used to fit the embedded distance pairs $d_{ij}(\mathbf{X})$. 

In essence, $\hat{d_{ij}}=\mathcal{F}(\delta_{ij})$ where $\mathcal{F}$ is usually an affine transformation\footnote{Monotone and polynomial regression transformations are employed for nonmetric-MDS, as well as, a wider family of transformations \cite{france2011two}.} $\hat{d_{ij}}= \alpha + \beta \delta_{ij}$ for unknown $\alpha$ and $\beta$.  Kruskal proposed an iterative gradient-based algorithm for the minimization of $\sigma_1$ since the solution cannot be expressed in closed form. 
% \geonote{which cannot be expressed in closed form: delete
% \thynote{no delete -- its vital for checking whether we need an iterative approach or not}
% \geonote{Ok I may have missed it. Is this referenced again later?}
% \thynote{Ok check now}
% }
Assuming that $\hat{d_{ij}=\delta_{ij}}$ the algorithm iteratively tries to find the coordinates of points $\mathbf{X}$ which are lying in the low embedding space $\mathbb{R}^L$. Trivial solutions ($\mathbf{X}=\mathbf{0}$ and $\hat{\mathbf{D}}=\mathbf{0}$) are avoided by the denominator term in Eq. \ref{eq stress1}. 

A weighted MDS raw Stress function is defined as: %in Eq.~\ref{eq raw_stress}, 
\begin{equation}
\label{eq raw_stress}
\sigma_{raw}^2(\mathbf{X},\hat{\mathbf{D}}) = 
\sum_{i=1}^{N} \sum_{j=1}^{N} w_{ij}(\hat{d_{ij}}-d_{ij}(\mathbf{X}))^2 
\end{equation}
\noindent
where the weights $w_{ij}$ are restricted to be non-negative; for missing data the weights are set equal to zero. 
%In this case, MDS does not take into account the effect of dissimilarity pair $\delta_{ij}$. 
By setting $w_{ij}=1, \space \forall 1 \leq i,j \leq N$ one can model an equal contribution to the Metric-MDS solution for all the elements.

\subsection{SMACOF}
SMACOF which stands for Scaling by Majorizing a Complex Function is a state-of-the-art algorithm for solving metric MDS and was introduced by \cite{Leeuw77applicationsof}. By setting $\hat{d_{ij}}=\delta_{ij}$ in raw stress function defined in Eq.~\ref{eq raw_stress}, SMACOF minimizes the resulting stress function $\sigma_{raw}^2(\mathbf{X})$.
%Eq.~\ref{eq smacof_stress}. 
\begin{equation}
\label{eq smacof_stress}
\sigma^2(\mathbf{X}) = 
\sum_{i=1}^{N} \sum_{j=1}^{N} w_{ij}(\delta_{ij}^2-2 \delta_{ij} d_{ij}(\mathbf{X}) + d_{ij}^2(\mathbf{X})) 
\end{equation}
The algorithm proceeds iteratively and decreases stress monotonically up to a fixed point by optimizing a convex function which serves as an upper bound for the non-convex stress function in Eq.~\ref{eq smacof_stress}. An extensive description of SMACOF can be found in \cite{borg_groenen_2005} while its convergence for a Euclidean embedded space $\mathbb{R}^L$ has been proven by \cite{deLeeuw1988}.

Let matrices $\mathbf{U}$ and $\mathbf{R(\mathbf{X})}$ be defined element-wise as follows: 
%in equations \ref{eq Moore Penrose} and \ref{eq supporting matrix} 

\begin{equation}
\label{eq Moore Penrose}
u_{ij} = 
\left\lbrace
\begin{array}{ll}
      -w_{ij} & i \ne j \\
      \sum_{k \ne i} w_{ik} & i = j
\end{array} 
\right.
\end{equation}

\begin{equation}
\label{eq supporting matrix}
r_{ij} = 
\left\lbrace
\begin{array}{ll}
      -w_{ij}\delta_{ij}d_{ij}^{-1}(\mathbf{X}) & i \ne j, d_{ij}(\mathbf{X}) \ne 0 \\
      0 & i \ne j, d_{ij}(\mathbf{X}) = 0 \\
      \sum_{k \ne i} r_{ik} & i = j
\end{array} 
\right.
\end{equation}

The stress function in Eq.~\ref{eq smacof_stress} is converted to the following quadratic form: 
\begin{equation}
\label{eq smacof_stress_converted}
\sigma^2(\mathbf{X}) = 
\sum_{i=1}^{N} \sum_{j=1}^{N} w_{ij}\delta_{ij}^2
- 2 tr(\mathbf{X}^T \mathbf{R}(\mathbf{X}) \mathbf{X})
+tr(\mathbf{X}^T \mathbf{U} \mathbf{X}) 
\end{equation}
%its equivalent form in Eq. \ref{eq smacof_stress_converted}.
The quadratic can be minimized iteratively as follows:
\begin{equation}
\label{eq majorizing func}
\begin{split}
T(\mathbf{X}, \hat{\mathbf{X}}^{(k)}) & = c - 2 tr(\mathbf{X}^T \mathbf{R}(\hat{\mathbf{X}}^{(k)}) \hat{\mathbf{X}}^{(k)}) +tr(\mathbf{X}^T \mathbf{U} \mathbf{X})  \\
& c = \sum_{i=1}^{N} \sum_{j=1}^{N} w_{ij}\delta_{ij}^2 = const.  
\end{split}
\end{equation}

\begin{equation}
\label{eq SMACOF update}
\hat{\mathbf{X}}^{(k+1)}=\underset{\mathbf{X}}{\operatorname{argmin}}\;{T(\mathbf{X}, \hat{\mathbf{X}}^{(k)})}=\mathbf{U}^{\dagger}\mathbf{R}(\hat{\mathbf{X}}^{(k)}) \hat{\mathbf{X}}^{(k)}
\end{equation}
where $\hat{\mathbf{X}}^{(k)}$ is the estimate of matrix $\mathbf{X}$ at the $k$th iteration and $\mathbf{U}^{\dagger}$ is Moore-Penrose pseudoinverse of $\mathbf{U}$.
At iteration $k$ the convex majorizing convex function touches the surface of $\sigma$ at the point $\hat{\mathbf{X}}^{(k)}$. By minimizing this simple quadratic function in Eq.~\ref{eq majorizing func} we find the next update which serves as a starting point for the next iteration $k+1$. The solution to the minimization problem is shown in Eq.~\ref{eq SMACOF update}. The algorithm stops when the new update yields a decrease $\sigma^2(\hat{\mathbf{X}}^{(k+1)})-\sigma^2(\hat{\mathbf{X}}^{(k)})$ that is smaller than a threshold value.

%\section{GENERAL PATTERN SEARCH (gps)\label{section:GPS}  }

\subsection{GPS formulation}
\label{section: GPS formulation}
The unconstrained problem of minimizing a continuously differentiable function $f:\mathbb{R}^n \rightarrow \mathbb{R}$ is formally described as %by Eq. \ref{eq General minimization}.

\begin{equation}
\label{eq General minimization}
\mathbf{x}^*=\underset{\mathbf{x} \in \mathbb{R}^n}{\operatorname{argmin}} \;f(\mathbf{x})
%{\min}f(\mathbf{x})
\end{equation}

\noindent
Next we present a short description of iterative GPS minimization of Eq.~\ref{eq General minimization} based on \cite{torczon1997convergence,dolan2003local}. First we have to define the following components:
\begin{itemize}
\item A basis matrix that could be any nonsingular matrix $\mathbf{B} \in \mathbb{R}^{n \times n}$. 
\item A matrix $\mathbf{C}^{(k)}$ for generating all the possible moves for the $k$th iteration  of the minimization algorithm %as defined in Eq.~\ref{eq GPS generating matrix}, 
\begin{equation}
\label{eq GPS generating matrix}
\mathbf{C}^{(k)}=[\mathbf{M}^{(k)} \enskip -\mathbf{M}^{(k)} \enskip \mathbf{L}^{(k)}] = [\mathbf{\Gamma}^{(k)} \enskip \mathbf{L}^{(k)}]
\end{equation}
\noindent
where the columns of $\mathbf{M}^{(k)} \in \mathbb{Z}^{n \times n}$  form a positive span of $\mathbb{R}^{n}$ and $\mathbf{L}^{(k)}$ contains at least the zero column of the search space $\mathbb{R}^{n}$.  
\item A pattern matrix $\mathbf{P}^{(k)}$ defined as
%in Eq. \ref{eq GPS pattern matrix} and 
\begin{equation}
\label{eq GPS pattern matrix}
\mathbf{P}^{(k)}=\mathbf{B}\mathbf{C}^{(k)}=[\mathbf{B}\mathbf{M}^{(k)} \enskip -\mathbf{B}\mathbf{M}^{(k)} \enskip \mathbf{B}\mathbf{L}^{(k)}]
\end{equation}
where the submatrix $\mathbf{B}\mathbf{M}^{(k)}$ forms a basis of $\mathbb{R}^{n}$. 
\end{itemize}

In each iteration $k$, we define a set of steps $\{\mathbf{s}_i^{(k)}\}_{i=1}^{m}$ generated by the pattern matrix $\mathbf{P}^{(k)}$ as shown next: 
% in Eq.~\ref{eq GPS trial step}.
\begin{equation}
\label{eq GPS trial step}
\mathbf{s}_i^{(k)}=\Delta^{(k)} \mathbf{p}_{i}^{(k)}, \enskip \mathbf{P}^{(k)} = [\mathbf{p}_{1}^{(k)},...,\mathbf{p}_{m}^{(k)}] \in \mathbb{R}^{n \times m}
\end{equation}
\noindent
where $\mathbf{p}_{i}^{(k)}$ is the $i$th column of $\mathbf{P}^{(k)}$ and defines the direction of the new step, while $\Delta^{(k)}$ configures the length towards this direction. If the pattern matrix $\mathbf{P}^{(k)}$ contains $m$ columns, then $m \geq n + 1$ in order to positively span the search space $\mathbb{R}^{n}$. Thus, a new trial point of GPS algorithm towards this step would be $\mathbf{x}_i^{(k+1)}=\mathbf{x}^{(k)}+\mathbf{s}_i^{(k)}$ where we evaluate the value of the function $f$ to minimize. The success of a new trial point is decided based on the condition that it takes a step towards further minimizing the function $f$, i.e., $f(\mathbf{x}^{(k)}+\mathbf{s}_i^{(k)}) > f(\mathbf{x}_i^{(k+1)})$. The steps of a GPS method are presented in Alg. \ref{alg: GPS algorithm}.

\begin{algorithm}[H]
\caption{General Pattern Search (GPS)}
\label{alg: GPS algorithm}
\begin{algorithmic}[1] 
% Where $x_0 \in \mathbb{R}^n$, $\Delta_0 > 0$
\Procedure{GPS\_SOLVER}{$\mathbf{x}^{(0)}$, $\Delta^{(0)}$, $\mathbf{C}^{(0)}$, $\mathbf{B}$}
\State $k=-1$
\Do
	\State $k = k+1$
% 	\State Compute $f(x_k)$
    \State $\mathbf{s}^{(k)} =$ EXPLORE\_MOVES($\mathbf{B}\mathbf{C}^{(k)}$, $\mathbf{x}^{(k)}$, $\Delta^{(k)}$) \label{func: GPS exploratory moves}
    \State $\rho^{(k)}=f(\mathbf{x}^{(k)}+\mathbf{s}^{(k)})-f(\mathbf{x}^{(k)})$
    \If{$\rho^{(k)} < 0$}
    	\State $\mathbf{x}^{(k+1)}=\mathbf{x}^{(k)} + \mathbf{s}^{(k)}$
        \Comment{Successful iteration}
    \Else 
    	\State $\mathbf{x}^{(k+1)}=\mathbf{x}^{(k)}$
        \Comment{Unsuccessful iteration}
    \EndIf
    \State $\Delta^{(k+1)}, \mathbf{C}^{(k+1)}=$ UPDATE($\mathbf{C}^{(k)}$, $\Delta^{(k)}$,  $\rho^{(k)}$) \label{func: GPS update ck and dk}
    
\doWhile{convergence criterion == False}
\EndProcedure
\end{algorithmic}
\end{algorithm}

To initialize the algorithm we select a point $\mathbf{x}^{(0)} \in \mathbb{R}^n$ and a positive step length parameter $\Delta^{(0)} > 0$. In each 
iteration $k$, we explore a set of moves defined by the $\texttt{EXPLORE\_MOVES}()$ subroutine at line~\ref{func: GPS exploratory moves} of the algorithm. %, defined in Alg.~\ref{alg: GPS algorithm}, line~\ref{func: GPS exploratory moves}. 
Pattern search methods
% \geonote{to the point on how these exploratory moves are selected: on the way the exploratory moves are selected
% or better: on the heuristics used for selection of exploratory moves
% \thynote{ OK check now \\
% on how these exploratory moves are selected. \\ $\Rightarrow$
% on the heuristics used for selection of exploratory moves }
% }
described using a GPS formalism mainly differ on the heuristics used for the selection of exploratory moves. If a new exploratory point lowers the value of the 
function $f$, iteration $k$ is successful and the starting point of the next iteration is updated $\mathbf{x}^{(k+1)}=\mathbf{x}^{(k)} + \mathbf{s}^{(k)}$ as shown in line~8, else there is no update. The step length parameter $\Delta^{(k)}$ is modified by the $\texttt{UPDATE}()$ subroutine  in line~11. For successful iterations, i.e.,  $\rho^{(k)} < 0$, the step length is forced to increase in a determistic way as follows:
\begin{equation}
\label{eq Delta Increase}
\begin{split}
& \Delta^{(k+1)} = \lambda^{(k)} \Delta^{(k)}, \enskip \lambda^{(k)}  \in \Lambda = \{\tau^{w_1},...,\tau^{w_{|\Lambda|}}\} \\
& \tau > 1, \enskip \{w_1,...,w_{|\Lambda|}\} \subset \mathbb{N}, \enskip |\Lambda| < +\infty 
\end{split}
\end{equation}
where $\tau$ and $w_i$ are predefined constants that are used for the $i$th successive successful iteration.
For unsuccessful iterations the step length parameter is decreased, i.e., $\Delta^{(k+1)} \leq \Delta^{(k)}$ as follows: %defined in Eq. \ref{eq Delta Decrease}.
\begin{equation}
\label{eq Delta Decrease}
\Delta^{(k+1)} = \theta \Delta^{(k)}, \enskip \theta = \tau^{w_0}, \enskip \tau > 1, \enskip w_0 < 0, 
\end{equation}
where $\tau$ and the negative integer $w_0$ determine the fixed ratio of step reduction.
Note that the generating matrix $\mathbf{C}^{(k+1)}$ could be also updated for unsuccessful/successful iterations in order to contain more/less search directions, respectively.

\subsection{GPS Convergence}
\label{section: GPS Convergence}
GPS methods under the aforementioned defined framework have some important convergence properties shown in \cite{torczon1997convergence, dolan2003local, doi:10.1137/S1052623496300507, doi:10.1137/0728030, doi:10.1137/S1052623497331373} and summarized here.
For any GPS method which satisfies the specifications of Hyp. \ref{hyp: mild exploratory moves} on the exploratory moves one may be able to show convergence for Alg.~\ref{alg: GPS algorithm}.
\begin{hyp}[Weak Hyp. on Exploratory Moves]
\label{hyp: mild exploratory moves} 
The subroutine\\ $\texttt{EXPLORE\_MOVES}()$ as defined in Alg.~\ref{alg: GPS algorithm}, line~\ref{func: GPS exploratory moves} guarantees the following:
\begin{itemize}
\item The exploratory step direction for iteration $k$ is selected from the columns of the pattern matrix $\mathbf{P}^{(k)}$ as defined in Eq.~\ref{eq GPS trial step} and the exploratory step length is ${\Delta^{(k)}}$ as defined in Eqs.~\ref{eq Delta Increase},~\ref{eq Delta Decrease}.
\item If among the exploratory moves $\mathbf{a}^{(k)}$ at iteration $k$ selected from the columns of the matrix $\Delta^{(k)}\mathbf{B}[\mathbf{M}^{(k)}  -\mathbf{M}^{(k)}]$ exist at least one move that leads to success, i.e., $f(\mathbf{x}^{(k)}+\mathbf{a}) < f(\mathbf{x}^{(k)})$, then the $\texttt{EXPLORE\_MOVES}()$  subroutine will return a move $\mathbf{s}^{(k)}$ such that 
$f(\mathbf{x}^{(k)}+\mathbf{s}^{(k)})<f(\mathbf{x}^{(k)})$.
%i.e., $f(\mathbf{x}^{(k)})$ decreases monotonically, i.e., $f(\mathbf{x}^{(k)}+\mathbf{s}^{(k)})<f(\mathbf{x}^{(k)})$, where $\mathbf{s}^{(k)} = \underset{\mathbf{a} \in \Delta^{(k)}[\mathbf{M}^{(k)}  -\mathbf{M}^{(k)}]}{\operatorname{argmin}} f(\mathbf{x}^{(k)}+\mathbf{a})$  
\end{itemize}
\end{hyp} 
Hyp. \ref{hyp: mild exploratory moves} enforces some mild constraints on the configuration of the exploratory moves produced by Alg. \ref{alg: GPS algorithm}, line~\ref{func: GPS exploratory moves}. Essentially, the suggested step $\mathbf{s}^{(k)}$ is derived from the pattern matrix $\mathbf{P}^{(k)}$, while the algorithm needs to provide a simple decrease for the objective function $f$. Specifically, the only way to accept an unsuccessful iteration would be if none of the steps from the columns of the matrix $\Delta^{(k)}\mathbf{B}[\mathbf{M}^{(k)}  -\mathbf{M}^{(k)}]$ lead to a decrease of the objective function $f$. Based on this hypothesis one can formulate 
Thm. \ref{theorem: Mild Convergence} as follows:

\begin{thm}
\label{theorem: Mild Convergence}
Let $L(\mathbf{x}^*)=\{\mathbf{x}: f(\mathbf{x}) \leq f(\mathbf{x}^*)\}$ be closed and bounded and $f$  continuously differentiable on a neighborhood of $L(\mathbf{x}^*)$, namely on the union of the open balls $\underset{{\mathbf{a} \in L(\mathbf{x}^*)}}{\bigcup} B(\mathbf{a}, \eta)$ where $\eta > 0$. If a GPS method is formulated as described in Section \ref{section: GPS formulation} and Hyp. \ref{hyp: mild exploratory moves} holds then for the sequence of iterations $\{\mathbf{x}^{(k)}\}$ produced by Alg.~\ref{alg: GPS algorithm} 
\[\underset{k \rightarrow +\infty}{\lim} \inf \; ||\nabla f(\mathbf{x}^{(k)})|| = 0 \]
\begin{proof}
See \cite{torczon1997convergence}.
\end{proof}
\end{thm}
% We simply state Thm. \ref{theorem: Mild Convergence} its detailed proof can be found in \cite{torczon1997convergence}. 
%and the function $f$ to be minimized enforces the criteria such that   holds then we are sure that: $\underset{k \rightarrow +\infty}{\lim} \inf \{||\nabla f(x_k)||\}$.
% \potamnote{Thm. \ref{theorem: Mild Convergence} provides a first order optimality condition if Hyp. \ref{hyp: mild exploratory moves} is true and the objective function $f$ is locally differentiable.
% \thynote{Check at the end of this section}}

As shown in \cite{Audet2004} one can construct a continuously differentiable objective function and a GPS method with infinite many limit points with non-zero gradients and thus even Thm. \ref{theorem: Mild Convergence} holds, the convergence of $||\nabla f(x_k)||$ is not assured. However, the convergence properties of GPS methods can be further strengthened if additional criteria are met. Specifically, a stronger hypothesis on exploratory moves Hyp. \ref{hyp: strong exploratory moves} regulates the measure of decrease of the objective function for each step produced by the GPS method, as follows:

\begin{hyp}[Strong Hyp. on Exploratory Moves]
\label{hyp: strong exploratory moves} 
The subroutine $\texttt{EXPLORE\_MOVES}()$ as defined in Alg.~\ref{alg: GPS algorithm}, line~\ref{func: GPS exploratory moves} guarantees the following:
\begin{itemize}
\item The exploratory step direction for iteration $k$ is selected from the columns of the pattern matrix $\mathbf{P}^{(k)}$ as defined in Eq.~\ref{eq GPS trial step} and the exploratory step length is ${\Delta^{(k)}}$ as defined in Eqs.~\ref{eq Delta Increase},~\ref{eq Delta Decrease}.
\item If among the exploratory moves $\mathbf{a}^{(k)}$ at iteration $k$ selected from the columns of the matrix $\Delta^{(k)}\mathbf{B}[\mathbf{M}^{(k)}  -\mathbf{M}^{(k)}]$ exists at least one move that leads to success, i.e., $f(\mathbf{x}^{(k)}+\mathbf{a}) < f(\mathbf{x}^{(k)})$, then the $\texttt{EXPLORE\_MOVES}()$  subroutine will return a move $\mathbf{s}^{(k)}$ such that:\\ 
$f(\mathbf{x}^{(k)}+\mathbf{s}^{(k)}) \leq \underset{\mathbf{a}^{(k)}}{\operatorname{min}} \; f(\mathbf{x}^{(k)} +\mathbf{a}^{(k)})$.
%i.e., $f(\mathbf{x}^{(k)})$ decreases monotonically, i.e., $f(\mathbf{x}^{(k)}+\mathbf{s}^{(k)})<f(\mathbf{x}^{(k)})$, where $\mathbf{s}^{(k)} = \underset{\mathbf{a} \in \Delta^{(k)}[\mathbf{M}^{(k)}  -\mathbf{M}^{(k)}]}{\operatorname{argmin}} f(\mathbf{x}^{(k)}+\mathbf{a})$  
\end{itemize}
\end{hyp} 

% \begin{hyp}[Strong Hyp. on Exploratory Moves]
% \label{hyp: strong exploratory moves} 
% The subroutine EXPLORE\_MOVES() as defined in algorithm \ref{alg: GPS algorithm}, line: \ref{func: GPS exploratory moves} should guarantee the 2 following things:
% \begin{itemize}
% \item New exploratory steps for $k$ iteration should be derived from the pattern matrix $P_k$ as defined in Eq.~\ref{eq GPS trial step}.
% \item If $min\{f(x_k+a), a \in \Delta^{(k)}[M_k  -M_k]\}$ then \\ $f(x_k+s_k) \leq min\{f(x_k+a), a \in \Delta^{(k)}[M_k  -M_k]$
% \end{itemize}
% \end{hyp} 
\noindent
Hyp. \ref{hyp: strong exploratory moves} enforces the additional strong constraint on the configuration of the exploratory moves, namely that the subroutine $\texttt{EXPLORE\_MOVES}()$ will do no worse than produce the best exploratory move from the columns of the matrix $\Delta^{(k)}\mathbf{B}[\mathbf{M}^{(k)}  -\mathbf{M}^{(k)}]$.
% Essentially, the exploratory step which is returned by the exploratory moves routine in algorithm \ref{alg: GPS algorithm}, line: \ref{func: GPS exploratory moves} should provide sufficient decrease of the objective function comparing to the individual directions defined in $\Delta^{(k)} \Gamma_k = \Delta^{(k)} [M_k -M_k]$.
Based on this hypothesis and by adding requirements restricting the exploration step direction and length for the GPS method, one can formulate Thm. \ref{theorem: Strong Convergence} which is also presented here without proof.
\begin{thm}
\label{theorem: Strong Convergence}
Let $L(\mathbf{x}^*)=\{\mathbf{x}: f(\mathbf{x}) \leq f(\mathbf{x}^*)\}$ be closed and bounded and $f$  continuously differentiable on a neighborhood of $L(\mathbf{x}^*)$, namely on the union of the open balls $\underset{{\mathbf{a} \in L(\mathbf{x}^*)}}{\bigcup} B(\mathbf{a}, \eta)$ where $\eta > 0$. If a GPS method is formulated as described in Section \ref{section: GPS formulation}, $\underset{k \rightarrow +\infty}{\lim} \Delta^{(k)} = 0$, the columns of the generating matrices $\mathbf{C}^{(k)}$ are bounded by norm and Hyp. \ref{hyp: strong exploratory moves} holds then for the sequence of iterations $\{\mathbf{x}^{(k)}\}$ produced by Alg.~\ref{alg: GPS algorithm}  
\[\underset{k \rightarrow +\infty}{\lim} \; ||\nabla f(\mathbf{x}^{(k)})|| = 0 \]
\begin{proof}
See \cite{torczon1997convergence}.
\end{proof}
\end{thm}
The additional requirements specify that: 1) the generating matrix $\mathbf{C}^{(k)}$ should be norm bounded in order to produce trial steps from Eq.~\ref{eq GPS trial step} that are bounded by the step length parameter $\Delta^{(k)}$ and 2) $\underset{k \rightarrow +\infty}{\lim} \Delta^{(k)} = 0$ that can be easily met by selecting  $\Lambda = \{1\}$ in  Eq.~\ref{eq Delta Decrease}; this also guarantees a non increasing sequence of $\Delta^{(k)}$ steps \cite{torczon1997convergence}.   
Although these criteria provide much stronger convergence properties, we are faced with a trade off between the theoretical proof of convergence and the efficiency of heuristics in finding a local optimum. 

Both theorems \ref{theorem: Mild Convergence} and \ref{theorem: Strong Convergence} provide a first order optimality condition if their specifications hold. Although the latter theorem premises much stronger convergence results, step-length control parameter $\Delta^{(k)}$, provides a reliable asymptotic measure of first-order stationarity when it is reduced after unsuccessful iterations \cite{dolan2003local}. 

% very useful review \href{http://thales.cheme.cmu.edu/dfo/comparison/dfo.pdf}{Derivative Free methods Review} Our algorithm cannot be completely reduced to Hooke Jeeves but to a more general family of patter search derivative-free algorithms for global optimization. To this end, further theoretical discussion and analysis should be done!
% \textit{Given $G = \{d(1), . . . , d(p)\}$ with $p \geq n + 1$ and $d
% (i) ∈ \mathbb{R}^n$, the function f is
% evaluated at a set of trial points $P_k = \{xk + \Delta^{(k)}d : d \in Gk\}$, where $\Delta^{(k)}$ is the
% step length. An iteration is successful if there exists y ∈ Pk such that $f(y) < f(xk) − \rho(\Delta^{(k)})$, where $\rho$ is a forcing function. The opportunistic poll strategy
% proceeds to the next iteration upon finding a point y, while the complete
% poll strategy evaluates all points in Pk and assigns $y = argmin_{x \in P_k}
% f(x)$.}

% The reduction to derivative methods and convergence properties by using rational lattices should be our guideline. This paper \cite{torczon1997convergence} provides insightful information and rigorous proofs towards this direction. Our algorithm is directly associated to General Pattern Search methods which are proven to be able to converge to local minima under some differentiable assumptions. VIRGINIA TORCZON \cite{dolan2003local} might also be quite useful.

% \subsection{Metropolis or Something related to approximating MDS (Optional)}

\section{Pattern Search MDS \label{section:OUR ALGORITHM}  }
\subsection{Core algorithm}
\label{s:core-alg}

The key idea behind the proposed algorithm is to treat MDS as a derivative-free problem, using a variant of general pattern search optimization to minimize a loss function. The input to pattern search MDS is a $N \times N$ target dissimilarity matrix $\mathbf{T}$ and the target dimension $L$ of the embedding space. An overview of the algorithm shown in Alg.~\ref{a:proposed} is presented next.

The initialization process of the algorithm consists of: 1) random sampling of $N$ points in the embedded space and construction of the matrix $\mathbf{X}^{(0)} = [ \mathbf{x}_1^{(0)}, \mathbf{x}_2^{(0)}, ..., \mathbf{x}_N^{(0)} ] \in \mathbb{R}^{N \times L}$, %\enote{ from a uniform distribution} 
2) computing the embedded space dissimilarity matrix $\mathbf{D}^{(0)}$,
where the element $d_{ij}^{(0)}$ is the Euclidean distance between vectors $\mathbf{x}_i^{(0)}$ and $\mathbf{x}_j^{(0)}$ of $\mathbf{X}^{(0)}$, and 3) computing the initial approximation error $e^{(0)} = f(\mathbf{T}, \mathbf{D}^{(0)})$, where $e$ is the element-wise mean squared error (MSE) between the two matrices. The functional $f$ that we attempt to minimize is the normalized square of the Frobenius norm of the matrix $\mathbf{T} - \mathbf{D}$, i.e.,  $f(\mathbf{T}, \mathbf{D}) = (1/N^2)||\mathbf{T} - \mathbf{D}||_{F}^2$. Equivalently one may express $f$ element-wise as follows:
\begin{equation}
 f(\mathbf{T}, \mathbf{D}) = \frac{1}{N^2} \sum_{i = 1}^{N}\sum_{j=1}^{N}(t_{ij} - d_{ij})^2, \enskip \; \mbox{where} \; \mathbf{T}, \mathbf{D}\in \mathbb{R}^{N \times N}
\end{equation}

\begin{algorithm}[H]
\caption{Proposed MDS}
\label{a:proposed}
\begin{algorithmic}[1] 

\Procedure{MDS}{$\mathbf{T}$, $L$, $r^{(0)}$}
    \State $k$ $\gets$ $0$
    \Comment{k is the number of epochs}
	\State $\mathbf{X}^{(k)}$ $\gets$ UNIFORM($N \times L$)
    \State $\mathbf{D}^{(k)}$ $\gets$ DISTANCE\_MATRIX($\mathbf{X}^{(k)}$)
    \State $e^{(k)}$ $\gets$ $f(\mathbf{T}, \mathbf{D}^{(k)})$
    \State $e^{(k-1)}$ $\gets$ $+\infty$
    \State $r^{(k)}$ $\gets$ $r^{(0)}$
    \While{$r^{(k)} > \delta$}
        \If{$e^{(k-1)} - e^{(k)} \leq \epsilon \cdot e^{(k)}$} \label{al:eps}
        	\State $r^{(k)}$ $\gets$ $\frac{r^{(k)}}{2}$
        \EndIf
    	\State $\mathbf{S}$ $\gets$ SEARCH\_DIRECTIONS($r^{(k)}$, $L$)
		\ForAll{$x \in \mathbf{X}^{(k)}$}
			\State $\mathbf{X^{*}}, e^{*}$ $\gets$ OPTIMAL\_MOVE($\mathbf{X}^{(k)}$,$x$,$\mathbf{S}$,$e^{(k)}$) \label{al:move_line}
			\State $e^{(k-1)} \gets e^{(k)}$
			\State $e^{(k)}$ $\gets$ $e^*$
        	\State $\mathbf{X}^{(k)}$ $\gets$ $\mathbf{X^*}$
        \EndFor
        \State $k = k + 1$
	\EndWhile
\EndProcedure
\end{algorithmic}
\end{algorithm}

\begin{figure}[h]
  \centering
  \includegraphics[width=0.3\textwidth]{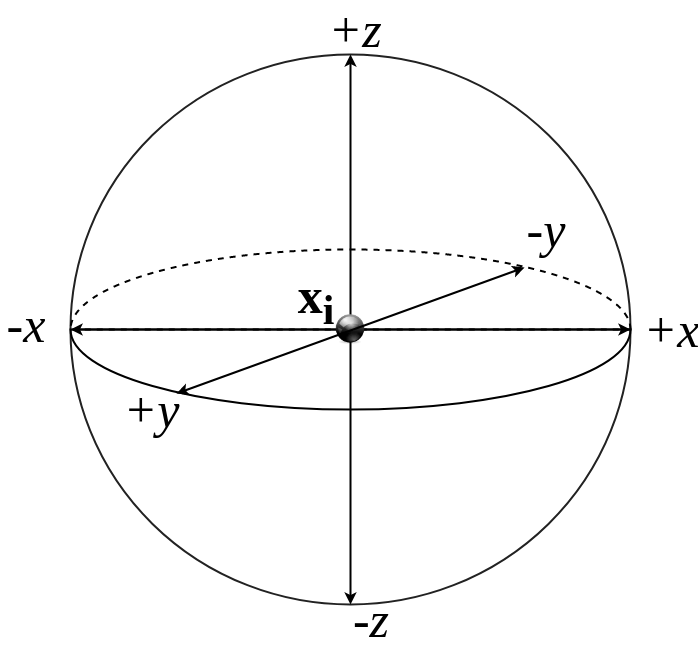}
\caption{Sphere of radius $r$ around point $\mathbf{x^{(k)}_i}$ and possible search directions}
\label{f:sphere}
\end{figure}

Following the initialization steps, in each epoch (iteration), we consider the surface of a hypersphere of radius $r$ around each point $\mathbf{x}^{(k)}_i$. The possible search directions lie on the surface of a hypersphere along the orthogonal basis of the space, e.g., in the case of $3$-dimensional space along the directions $\pm x,\pm y,\pm z$ on the sphere shown in Fig.~\ref{f:sphere}. This creates the search directions matrix $S$ and is summarized in Alg.~\ref{a:search_directions}

\begin{algorithm}[H]
\caption{Define search directions}
\label{a:search_directions}
\begin{algorithmic}[1] 

\Function{SEARCH\_DIRECTIONS}{$r$, $L$}
	\State $\mathbf{S^+}$ $\gets$ $r \cdot \mathbf{I}_{L }$
	\State $\mathbf{S^-}$ $\gets$ $-r \cdot \mathbf{I}_{L }$
	\State $\mathbf{S}$ $\gets$ $[\frac{\mathbf{S^+}}{\mathbf{S^-}}]$
    \State \Return $S$
\EndFunction
\end{algorithmic}
\end{algorithm}

Each point is moved greedily %, without considering other points, 
along the dimension that produces the minimum error.
At this stage we only consider moves that yield a monotonic decrease in the error function. Alg. \ref{a:move} finds the optimal move that minimizes $e^{(k)} = f(\mathbf{T},\mathbf{D}^{(k)})$ for each new point $\tilde{x}$ and moves $\mathbf{X}$ in that direction. Note that when writing $s \in \mathbf{S}$, the matrix $\mathbf{S}$ is considered to be a set of row vectors.

\begin{algorithm}[H]
\caption{Find optimal move for a point}
\label{a:move}
\begin{algorithmic}[1] 

\Function{OPTIMAL\_MOVE}{$\mathbf{X}^{(k)}$, $x$, $S$, $e$}
	\State $e^* \gets e$
	\ForAll{$s \in \mathbf{S}$}
		\State $\tilde{x}$ $\gets$ $x + s$
    	\State $\mathbf{\tilde{X}}$ $\gets$ UPDATE\_POINT($\mathbf{X}^{(k)}$, $x$, $\tilde{x}$)
    	\Comment{Update $x$ point of $\mathbf{X}^{(k)}$ with $\tilde{x}$}
    	\State $\mathbf{D}$ $\gets$ DISTANCE\_MATRIX($\mathbf{\tilde{X}}$) \label{al:distance_matrix}
    	\State $\tilde{e}$ $\gets$ f($\mathbf{T}$, $\mathbf{D}$)
    	\If{$\tilde{e} < e^*$}
    		\State $e^*$ $\gets$ $\tilde{e}$
    		\State $\mathbf{X}^*$ $\gets$ $\mathbf{\tilde{X}}$
   		\EndIf
	\EndFor
    \State \Return $X^*, e^*$
\EndFunction
\end{algorithmic}
\end{algorithm}

%\potamnote{This is not what you have in Alg.2 - looks like a bug - right now you update the radius after each point is moved not after all points are moved}

The resulting error $e^*$ is computed after performing the optimal move for each point in  $\mathbf{X}^{(k)}$. If the error decrease hits a plateau, we halve the search radius and proceed to the next epoch. This is expressed as $e^{(k)} - e^* < \epsilon \cdot e^{(k)}$, where $\epsilon$ is a small positive constant, namely the error decrease becomes very small in relation to $e^{(k)}$. The process stops when the search radius $r$ becomes very small, namely $r < \delta$, where $\delta$ is a small constant, as shown in  Alg.~\ref{a:proposed}.

\subsection{Optimizations and algorithm complexity}
\label{section: Optimizations and algorithm complexity}
Next, a set of algorithmic optimizations are presented that can improve the execution time and the solution quality of Alg.~\ref{a:proposed}. We also present ways to improve the execution time by searching for an approximate solution, as well as, discuss ways to utilize parallel computation for parts of the algorithm.

\subsubsection{Allow for ``bad'' moves}
\label{section: bad moves}
In Section \ref{s:core-alg} we restrict the accepted moves so that the error decreases monotonically. This is a reasonable restriction that also provides us with theoretical guarantees of convergence. Nonetheless in our experimental setting, we observed that if we relax this restriction and allow each point to always make the optimal move, regardless if the error (temporarily) increases the algorithm converges faster to better solutions. The idea of allowing greedy algorithms to make some ``bad'' moves in hope to get over local minima can be found in other optimization algorithms, simulated annealing \cite{kirkpatrick1983optimization} being the most popular. To implement this one can modify line~\ref{al:move_line} in Alg.~\ref{a:proposed} to:

\begin{algorithm}[H]
\begin{algorithmic}
\State $\mathbf{X^{*}}, e^{*}$ $\gets$ OPTIMAL\_MOVE($\mathbf{X}^{(k)}$, $x$, $S$, $+\infty$)
\end{algorithmic}
\end{algorithm}

\subsubsection{Online computation of dissimilarity matrix}

In line~\ref{al:distance_matrix} of Alg.~\ref{a:move} we observe that we recompute the dissimilarity matrix for each move. This can be avoided because each move modifies only one point $\mathbf{x}^{(k)}_{i}$, therefore only the row $\mathbf{d}^{(k)}_{i,:}$ and column $\mathbf{d}^{(k)}_{:,i}$ of the dissimilarity matrix $\mathbf{D}^{(k)}$ are affected. Furthermore only one dimension $l$ of the vector $\mathbf{x}^{(k)}_{i}$ is modified by the move, i.e., only element $x^{(k)}_{i,l}$ of matrix $\mathbf{X}^{(k)}$.
In detail, the element $d_{i,j}$ that stores the dissimilarity between points $\mathbf{x}_{i}$ and $\mathbf{x}_{j}$ should be updated as follows for the move from $x_{i,l}^{(k)}$ to $x_{i,l}^{(k+1)}$ for $i \neq j$:
% . Also let $x_d$ be the dimension of point $x$ along which the point is moved by $r$, with $\tilde{x}_d = x_d \pm r$. Then the update of the elements of the dissimilarity matrix
% is given by the function in Eq. \ref{e:dm_update}.

\begingroup
\footnotesize
\begin{equation}
\label{e:dm_update}
d_{i,j}^{(k+1)} = \sqrt {\big(d_{i,j}^{(k)}\big)^2 - \big(x_{i,l}^{(k)} - x_{j,l}^{(k)}\big)^2 +\big(x_{i,l}^{(k+1)} - x_{j,l}^{(k+1)}\big)^2 }
%     u(D, x_d, \tilde{x}_d, y_d) = 
% \begin{cases}
% 	\sqrt{(D_{xy}^{k})^2 - (x_{d} - y_{d})^2 + (\tilde{x}_d - y_{d})^2},& x \neq y \\
%     0,              & x = y
% \end{cases}
\end{equation}
\endgroup \\

% and we can modify Alg. \ref{a:move} to Alg. \ref{a:move_opt}:

% \begin{algorithm}[H]
% \caption{Find optimal move for a point}
% \label{a:move_opt}
% \begin{algorithmic}[1] 

% \Function{OPTIMAL\_MOVE}{$D_k$, $\mathbf{X_k}$, $x$, $S$, $e$}
% 	\State $e^* \gets e$
%     \State $D \gets D_k$
% 	\ForAll{$s \in S$}
% 		\State $\tilde{x}$ $\gets$ $x + s$
%     	\State $\mathbf{\tilde{X}}$ $\gets$ UPDATE\_POINT($\mathbf{X_k}$, $x$, $\tilde{x}$)
%     	\Comment{Update $x$ point of $\mathbf{X_k}$ with $\tilde{x}$}
        
%         \State $d$ $\gets$ NON\_ZERO($s$)
%         \Comment{Modified dimension $d$ is the single non zero element of $s$}
%     	\ForAll{$y \in \{ \mathbf{X_k} \smallsetminus x \}$}
%         	\State $D_{xy} \gets u(D_k, x_d, \tilde{x_d}, y_d)$ 
%         \EndFor
%     	\State $\tilde{e}$ $\gets$ f($\Delta$, $D$)
%     	\If{$\tilde{e} < e^*$}
%     		\State $e_*$ $\gets$ $\tilde{e}$
%     		\State $\mathbf{X^*}$ $\gets$ $\mathbf{\tilde{X}}$
%            	\State $D_k$ $\gets$ $D$
%    		\EndIf
%         \State $D$ $\gets$ $D_k$
% 	\EndFor
%     \State \Return $X^*, e^*, D_k$
% \EndFunction
% \end{algorithmic}
% \end{algorithm}

\subsubsection{Step and move selection} %Approximation techniques}

%\begin{itemize}
% \item \potamnote{I think the first bullet sounds like parameter optimization that should be in the experimental procedure section? Also you might want to generalize the update here to what Thy uses in Alg.1, which includes both the step change AND the number of moves.} The constant $\epsilon$ in line~\ref{al:eps} of Alg.~ \ref{a:proposed} determines for how long moves are constrained on a search area of radius $r$. By setting $\epsilon$ to a value very close to $0$, e.g., $10^{-10}$, the search will take more epochs but the solution will be closer to the local optimum. If we relax $\epsilon$ to a value like $10^{-2}$, we can do a coarse exploration of the search space that will produce a rough solution in a small number of epochs. In our experiments we set $\epsilon=10^{-4}$ that provides a good trade-off between solution quality and fast convergence.

It follows from the need to search for the optimal move across the embedding dimensions $L$, that the complexity of the algorithm has a linear dependency on $L$. A large value of $L$ might affect the execution time of the algorithm. An approximate technique to alleviate this is perform a random sampling over all possible directions in the $L$ dimensional space in order to select a ``good'' direction instead of the optimal, thus restricting the search space\footnote{One can potentially do better than random sampling of all possible directions in the $L$ dimensional space. As the geometry of the embedding space starts becoming apparent, after a few epochs of the algorithm, it makes sense to increasingly bias the search towards the principal component vectors of the neighborhood of the point that is being moved.}. 

%\subsubsection{Automatic tuning of starting radius}

%\item 
An important parameter for our algorithm is the starting radius $r^{(0)}$. This parameter controls how broad the search will be initially and has an effect similar to the learning rate of gradient-based optimization algorithms. If we are too conservative and choose a small initial radius, the algorithm will converge slowly to a local optimum, whereas if we set it too high, the error will overshoot and convergence is not guaranteed. A simple technique to automatically find a good starting radius is to use binary search. In particular, we set the starting radius to an arbitrary value, perform a dry run of the algorithm for one epoch and observe the effect on error. If the error increases we halve the radius. Otherwise we double it and repeat the process. This process is allowed to run for a small number of epochs. The starting radius found using this technique is a not too pessimistic or too optimistic estimate of the best parameter value.
%\end{itemize}

\subsubsection{Parallelization}

Another way to boost the execution time is to utilize parallel computation to speed up parts of the algorithm. In our case we can parallelize the search for the optimal moves across the embedding dimensions using the map-reduce parallelization pattern. Specifically, we can map the search for candidate moves to run in different threads and store the error for each candidate move in an array 
$\mathbf{e} = [e_{1}, e_{2} .., e_{2L}]$. After the search completes we can perform a reduction operation (min) to find the optimal move and the optimal error $\mathbf{X}^*, e^*$. For our implementation we used the OpenMP parallelization framework \cite{dagum1998openmp} and it led to a $2-4$ times speedup in execution time.

\subsubsection{Complexity}

For each epoch we search across $2L$ dimensions for $N$ points. In each search we also need $\mathcal{O}(N)$ operations to update the distance matrix. Thus, the per epoch computational complexity of the algorithm is $\mathcal{O}(N^2L)$. The optimizations proposed above do not change the complexity of the algorithm per epoch with the notable exception of the move selection optimization:  if instead of 2$L$ moves per epoch one would consider only 2$K$ moves. In this case, the overall complexity per epoch would be $\mathcal{O}(N^2K)$ instead of $\mathcal{O}(N^2L)$. However, as we shall see in the experiments that follow the (rest of the) proposed optimization significantly improve convergence speed, resulting in fewer epochs and less computation complexity overall.

% ============= START Our algorithm as a GPS method START ==================

\subsection{GPS formulation of our Algorithm}
\label{section: Our Algorithm is a GPS}

Pattern Search MDS belongs to the general class of GPS methods and can be expressed using the unified GPS formulation introduced in Section \ref{section: GPS formulation}. Next, we express our proposed algorithm and associated objective function under this formalism. 

First, we restate the problem of MDS in a vectorized form. We  use matrix $\mathbf{\Delta}$ with elements $\{\delta_{ij}\}_{1 \leq i, j \leq N}$ that expresses the dissimilarities between $N$ points in the high dimensional space. The set of points $\{\mathbf{x}_i\}_{i=1}^N$ lie on the low dimensional manifold $\mathcal{M} \in \mathbb{R}^L$ and form the column set of matrix $\mathbf{X}^T$. The matrix $\mathbf{X} \in \mathbb{R}^{N \times L}$ will be now vectorized as an one column vector as shown next: 

\begin{equation}
\label{eq Vectorized Variable}
\begin{gathered}
\mathbf{x}_i = [x_{i1},...,x_{iL}]^T \in \mathbb{R}^{L} , 1 \leq i \leq N \\
\mathbf{z} = vec(\mathbf{X}^T) = [x_{11},...,x_{1L},...,x_{N1},...,x_{NL}]^T
\end{gathered}
\end{equation}

Now our new variable $\mathbf{z}$ lies in the search space $\mathbb{R}^{N \cdot L}$. The distance between any two points $\mathbf{x}_i$ and $\mathbf{x}_j$ of the manifold $\mathcal{M}$ remains the same but is now expressed as a function of the vectorized variable $\mathbf{z}$. Namely, $d_{ij}(\mathbf{X})=||\mathbf{x}_i - \mathbf{x}_j||= \sqrt{\sum_{k=1}^L (x_{ik}-x_{jk})^2}=d_{ij}(\mathbf{z})$. To this end, our new objective function to minimize $g$ is the MSE between the given dissimilarities $\delta_{ij}$ and the euclidean distances $d_{ij}$ in the low dimensional manifold $\mathcal{M}$ as defined in Eq. \ref{eq MSE of the vectorized var} shown next:
\begin{equation}
\label{eq MSE of the vectorized var}
g(\mathbf{z}) = \frac{1}{N^2} \sum_{i = 1}^{N}\sum_{j=1}^{N}(d_{ij}(\mathbf{z}) - \delta_{ij})^2, \enskip \mathbf{z} \in \mathbb{R}^{N \cdot L}
\end{equation}

Consequently, the initial MDS is now expressed as an unconstrained non-convex optimization problem which is expressed by minimizing the function $g$ over the search space of $\mathbb{R}^{N \cdot L}$ (Eq. \ref{eq Our algorithm's minimization}). Specifically, the $L$ coordinates for all $N$ points on the manifold $\mathcal{M}$ now serve as degrees of freedom for our solution.  

\begin{equation}
\label{eq Our algorithm's minimization}
\mathbf{z}^*=\underset{\mathbf{z} \in \mathbb{R}^{N \cdot L}}{\min}g(\mathbf{z})
\end{equation}

Now that we have formulated the problem and the variable $\mathbf{z}$ in the appropriate format we can match each epoch of our initial algorithm with an iteration of a GPS method. Therefore, the moves produced by our algorithm form a sequence of points $\{\mathbf{z}^{(k)}\}$. Moreover, we are going to define the matrices $\mathbf{B}, \mathbf{C}^{(k)}, \mathbf{P}^{(k)}$ for our algorithm as in Eqs.~\ref{eq GPS generating matrix}, \ref{eq GPS pattern matrix}. The choice of our basis matrix $\mathbf{B}$ is the identity matrix as shown in Eq.~\ref{eq Our algorithm's B}. 

\begin{equation}
\label{eq one hot vector}
\mathbf{e}_i = \underset{index \enskip i}{[0,..,\underbrace{1},...,0]}^T, 1 \leq i \leq N \cdot L 
\end{equation}

\begin{equation}
\label{eq Our algorithm's B}
\mathbf{B} = \mathbf{I}_{N \cdot L} = [\mathbf{e}_1,...,\mathbf{e}_{N \cdot L}]
\end{equation}

While the identity matrix is non singular and its columns span positively the search space $\mathbb{R}^{N \cdot L}$, we also define $\mathbf{M}^{(k)}$ as the identity matrix. In Eq. \ref{eq Our algorithm's Gammak} matrix 
$\mathbf{\Gamma}^{(k)}$ represents the movement alongside the unit coordinate vectors of $\mathbb{R}^{N \cdot L}$. Nevertheless, our generating matrix $\hat{\mathbf{C}}$ also comprises of all the remaining possible directions which are generated by the set $\{-1,0,1\}$. In total, we have $3^{N \cdot L}-2 \cdot N \cdot L$ extra direction vectors inside the corresponding matrix $\mathbf{L}^{(k)}$ as it is shown in Eq. \ref{eq Our algorithm's Lambdak}.

\begin{equation}
\label{eq Our algorithm's Gammak}
\begin{gathered}
\mathbf{M}^{(k)} = \hat{\mathbf{M}} = \mathbf{I}_{N \cdot L} \in \mathbb{Z}^{N \cdot L \times N \cdot L}\\
\mathbf{\Gamma}^{(k)} = \hat{\mathbf{\Gamma}} = [\hat{\mathbf{M}} \enskip -\hat{\mathbf{M}}]
\end{gathered}
\end{equation}

\begin{equation}
\label{eq Our algorithm's Lambdak}
\begin{gathered}
\hat{S}=\{-1,0,1\} \\
\mathbf{L}^{(k)} = \hat{\mathbf{L}} \\
\hat{\mathbf{L}} =\{\hat{v}: \hat{v} \in \underset{N \cdot L}{\underbrace{\hat{S} \times ... \times \hat{S}}} \land \hat{v} \notin \{\mathbf{e}_1,...,\mathbf{e}_{N \cdot L}\}\}
\end{gathered}
\end{equation}

According to Eqs.~\ref{eq Our algorithm's Gammak}, \ref{eq Our algorithm's Lambdak}, we construct the full pattern matrix $\mathbf{P}^{(k)}$ in Eq.~\ref{eq Our algorithm's Pk} in a similar way to Eq.~\ref{eq GPS pattern matrix}. For our algorithm the pattern matrix is is equal to our generating matrix $\mathbf{C}^{(k)} = \hat{\mathbf{C}}$ which is also fixed for all iterations. Conceptually, the generating matrix $\hat{\mathbf{C}}$ contains all the possible exploratory moves while a heuristic is utilized for evaluating the objective function $g$ only for a subset of them.

\begin{equation}
\label{eq Our algorithm's Pk}
\begin{gathered}
\mathbf{C}^{(k)}=\hat{\mathbf{C}}=[\hat{\mathbf{\Gamma}} \enskip \hat{\mathbf{L}}]=[\hat{\mathbf{M}} \enskip - \hat{\mathbf{M}} \enskip \hat{\mathbf{L}}] \\
\mathbf{P}^{(k)} = \hat{\mathbf{P}} \equiv \mathbf{B}\hat{\mathbf{C}} \equiv \hat{\mathbf{C}}
\end{gathered}
\end{equation}

Finally, we configure the updates of the step length parameter for each class of both successful and unsuccessful iterations as they were previously described in Eqs.~\ref{eq Delta Increase},~\ref{eq Delta Decrease}, respectively. Recalling the notation of Section \ref{section: GPS formulation}, $\hat{\mathbf{s}}^{(k)}$ is the step which is returned from our exploratory moves subroutine at $k$th iteration. For the successful iterates $g(\mathbf{z}^{(k)}+\hat{\mathbf{s}}^{(k)})<g(\mathbf{z}^{(k)})$ we do not further increase the length of our moves by limiting $\Lambda = \{1\}$ as follows:  

\begin{equation}
\label{eq Our algorithm's Delta Increase}
\Delta^{(k+1)} = \Delta^{(k)}, \enskip \; \mbox{if} \; \enskip f(\mathbf{z}^{(k)}+\hat{\mathbf{s}}^{(k)})<f(\mathbf{z}^{(k)}) 
\end{equation}

Similarly, for the unsuccessful iterations $g(\mathbf{z}^{(k)}+\hat{\mathbf{s}}^{(k)}) \geq g(\mathbf{z}^{(k)})$ we halve the distance by a factor of $2$ by setting $\theta = \frac{1}{2}$ as it is shown next:

\begin{equation}
\label{eq Our algorithm's Delta Decrease}
\Delta^{(k+1)} =  \frac{1}{2} \Delta^{(k)}, \enskip \; \mbox{if} \; \enskip f(\mathbf{z}^{(k)}+\hat{\mathbf{s}}^{(k)})>=f(\mathbf{z}^{(k)}) 
\end{equation}

A short description of our algorithm as a GPS method for solving the problem stated in Eq.~\ref{eq Our algorithm's minimization} follows: In each iteration, we fix the optimal coordinate direction for each one of the points lying on the low dimensional manifold $\mathbf{x}_i \in \mathcal{M}, \enskip 1 \leq i \leq N$. For each internal iteration of Alg.~\ref{a:move}, if the optimal direction produces a lower value for our objective function $g$ we accumulate this direction and move alongside this coordinate of the $\mathbb{R}^{N \cdot L}$. Otherwise, we remain at the same position. %The position that was set from the aforementioned procedure. 
As a result, the exploration of coordinates for the new point $\mathbf{x}_{i+1}$ begins from this temporary position. This greedy approach provides a potential one-hot vector as described in Eq.~\ref{eq one hot vector} if the iterate is successful or otherwise, the zero vector $\mathbf{0} \in \mathbb{R}^{N \cdot L}$. The final direction vector $\hat{\mathbf{s}}^{(k)}$ for $k$th iteration is computed by summing these one-hot or zero vectors. At the $k$th iteration, the movement would be given by a scalar multiplication of the step length parameter $\Delta^{(k)}$ with the final direction vector in a similar way as defined in Eq.~\ref{eq GPS trial step}. This provides a simple decrease for the objective function $g$ or in the worst case represent a zero movement in the search space $\mathbb{R}_{N \cdot L}$. Regarding the movement across $\hat{\mathbf{s}}^{(k)}$, it is trivial to show that this reduction of the objective function $g$ is an associative operation. In other words, accumulating all best coordinate steps for each point $\{\mathbf{x}_i\}_{i=1}^N$ and performing the movement at the end of the $k$th iteration (as GPS method formulation requires) produces the same result as taking each coordinate step individually. Finally, pattern search MDS terminates when the step length parameter $\Delta^{(k)}$ becomes smaller than a predefined threshold.

% $g(\mathbf{z}^{(k+1)}) = g(\mathbf{z}^{(k)} + \hat{\mathbf{s}}^{(k)}_i) \leq g(\mathbf{z}^{(k)})$.

\subsection{Convergence of our Algorithm}
Now that we have homogenized the notation framework as well as have expressed the proposed algorithm as a GPS method one can utilize the theorems stated in Section \ref{section: GPS Convergence}  to prove the convergence properties of the proposed algorithm. 

First of all, the objective function $g$ is indeed continuously differentiable for all the values of the search space $\mathbb{R}^{N \cdot L}$ by its definition in Eq. \ref{eq MSE of the vectorized var}. Moreover, the pattern matrix $\hat{\mathbf{P}}$ in Eq.~\ref{eq Our algorithm's Pk} contains all the possible step vectors provided by our exploratory moves routine. Thus, all of our exploratory moves are defined by Eq.~\ref{eq GPS trial step}. In each iteration we evaluate the trial steps alongside all coordinates for all the points $\mathbf{x}_i \in \mathcal{M}, \enskip 1 \leq i \leq N$. In our restated problem definition (see Section \ref{section: Our Algorithm is a GPS}), this is translated to searching all over the identity matrices $\mathbf{I}_{N \cdot L}$ and $-\mathbf{I}_{N \cdot L}$ of the search space $\mathbb{R}^{N \cdot L}$. But from our definition of the first columns of our generating matrix in Eq.~\ref{eq Our algorithm's Gammak} this corresponds to checking all the potential coordinate steps provided by $\hat{\mathbf{\Gamma}} = [\mathbf{I}_{N \cdot L} \enskip -\mathbf{I}_{N \cdot L}]$. Consequently, if there exists a simple decrease when moving towards any of the directions provided by the columns of $\hat{\mathbf{\Gamma}}$ then our algorithm also provides a simple decrease. This result verifies that Hyp.~\ref{hyp: mild exploratory moves} is true for the exploratory moves. By combining the differentiability of our objective function $g$ and  Hyp.~\ref{hyp: mild exploratory moves}, Thm.~\ref{theorem: Mild Convergence} holds for pattern search MDS. Hence, $\underset{k \rightarrow +\infty}{\lim} \inf \; ||\nabla f(\mathbf{z}^{(k)})|| = 0$ is guaranteed.

% \potamnote{do we really need to take the next step here ... do we have infinite vanishing gradients here?
% \thynote{The limit of the infimum of gradients converges to zero but this does not mean that the actual gradient goes to zro in some extreme cases. \\
% --
% As Audet showed in \cite{Audet2004} one can construct a continuously differentiable objective function and a GPS method with infinite many limit points with non-zero gradients and thus even Thm. \ref{theorem: Mild Convergence} holds, the convergence of $||\nabla f(x_k)||$ is not assured.
% --}}

Trying to further strengthen the convergence properties of the proposed algorithm,
we note that most of the requirements of Thm.~\ref{theorem: Strong Convergence} are met but we fail to meet the specifications of Hyp.~\ref{hyp: strong exploratory moves} for the minimum decrease provided by the the columns of $\hat{\mathbf{\Gamma}}$. However, our generating matrix $\hat{\mathbf{C}} = [\hat{\mathbf{c}}_1,...,\hat{\mathbf{c}}_{3^{N \cdot L}}]$ is indeed bounded by norm because $||\hat{\mathbf{c}}_j||_1 \leq N \cdot L, \enskip 1 \leq j \leq 3^{N \cdot L}$. By halving the step length parameter for the unsuccessful iterations we also ensure that $\lim_{k \rightarrow \infty} \Delta^{(k)}$. In order to meet the specifications of Thm.~\ref{theorem: Strong Convergence} we would need a quadratic complexity of $\mathcal{O}((N \cdot L)^2)$ in order to ensure that each iteration provides the same decrease in function $g$ as the decrease provided by the ``best'' column of $\hat{\mathbf{\Gamma}}$. This is formally stated at the second part of Hyp.~\ref{hyp: strong exploratory moves}. If we modify our algorithm in order to meet these requirements we would not be able to implement all the optimizations proposed in Section \ref{section: Optimizations and algorithm complexity} and the overall runtime would be dramatically increased.

% \potamnote{How does allowing for bad moves affect convergence. A detailed disussion of all the optimization should be added here ...
% \thynote{You mean another subsection, I guess... But why should we do this inside the section where we try to incorporate GPS notation for our algorithm?}
% \geonote{Also not sure how to do this theoretically since the optimizations are derived empirically.}
% }

% ============= END Our algorithm as a GPS method END ==================

\section{EXPERIMENTS\label{section:EXPERIMENTS}}

\subsection{Tuning the hyperparameters}
\label{ss:hyperparam}
\begin{figure}[htb!]
  \centering
  \includegraphics[width=.7\textwidth]{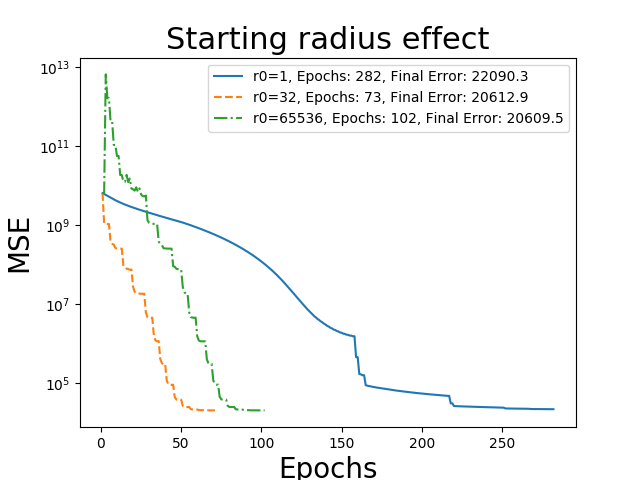}
\caption{Convergence plot for different starting radii}
\label{f:starting_radius}
\end{figure}

Next we present some guidelines on how to set the hyperparameters for the proposed algorithm and report the values used in the experiments that follow. Specifically:
\begin{itemize}
\item The constant $\epsilon$ in line~\ref{al:eps} of Alg.~\ref{a:proposed} determines when the move radius $r$ is decreased. By setting $\epsilon$ to a value very close to $0$, e.g., $10^{-10}$, the search will take more epochs but the solution will be closer to the local optimum. If we relax $\epsilon$ to a value around $10^{-2}$, we can do a coarse exploration of the search space that will produce a rough solution in a small number of epochs. In our experiments we set $\epsilon=10^{-4}$ that provides a good trade-off between solution quality and fast convergence for the datasets used.
\item 
%\potamnote{ From now on again this is part of the experimental procedure. How is 50\% affected by $L$ and a few other parameters?} 
We experimentally found that if $L$ is large, we may only search $50\%$ of the search dimensions and still get a good solution, while significantly reducing the execution time. For this to hold, it is important that we randomly sample a new search space for each epoch.
\item 
%\potamnote{this again should be part of the experimental section. Overall this sounds a bit naive compared to the algorithm in 1. Also there is some overlap with the discussion on $\epsilon$ above ... Need to review and decide how best to present ...} 
The proposed algorithm is relatively robust to the choice  of the initial size of the move search radius. However, the choice of $r^{(0)}$ does affect convergence speed.
We show the convergence for an example run of the classical swissroll (see Section \ref{ss:geometry}) for best-case ($r^{(0)}=32$), pessimistic ($r^{(0)}=1$) and optimistic ($r^{(0)}=65536$) starting radii in Fig.~\ref{f:starting_radius}.
\end{itemize}

\subsection{Manifold Geometry}
\label{ss:geometry}

The key assumption in manifold learning is that input data lie on a low-dimensional, non-linear manifold, embedded in a high-dimensional space. Thus non-linear dimensionality reduction techniques aim to extract the low-dimensional manifold from the high dimensional space. To showcase this we generated a variety of geometric manifold shapes and compared the proposed MDS to other, well-established dimensionality reduction techniques. We make the code to generate the synthetic data openly available to the community\footnote{Open source code available: \href{https://github.com/georgepar/gentlemandata}{https://github.com/georgepar/gentlemandata}}.

One should note that MDS algorithms with Euclidean distance matrices as inputs cannot infer data geometry, thus we need to provide as input a \textit{geodesic distance matrix}. This matrix is computed by running Djikstra's shortest path algorithm on the Nearest Neighbors graph trained on the input data. 
For our experiments we sample $3000$ points on $11$ $3D$ shapes and reduce them to $2$ dimensions using pattern search MDS, SMACOF MDS \cite{Leeuw77applicationsof}, truncated SVD \cite{golub1965calculating}, Isomap \cite{tenenbaum_global_2000,bernstein_graph_2000,zha_isometric_2003,DonohoG05,pless_image_2003}, Local Linear Embedding (LLE) \cite{BelkinN03,Cayton,saul_think_2003,sha_analysis_2005, belkin_laplacian_2001}, Hessian LLE \cite{zhang_principal_2004,donoho_hessian_2003}, modified LLE \cite{zhang2007mlle} and Local Tangent Space Alignment (LTSA) \cite{LTSA}. 

The geodesic distance matrices provided to pattern search MDS and SMACOF MDS is computed using Djikstra's algorithm on k-NN (nearest neighbor) graphs.
%with $k=20$ neighbors for all experiments.
%\thynote{Why 20? Default values look better}
We list the times it took each method to run. Note that pattern search MDS is faster than SMACOF MDS.

\begin{figure*}
\centering

\begin{subfigure}[b]{.48\textwidth}
  \centering
  \includegraphics[width=\linewidth]{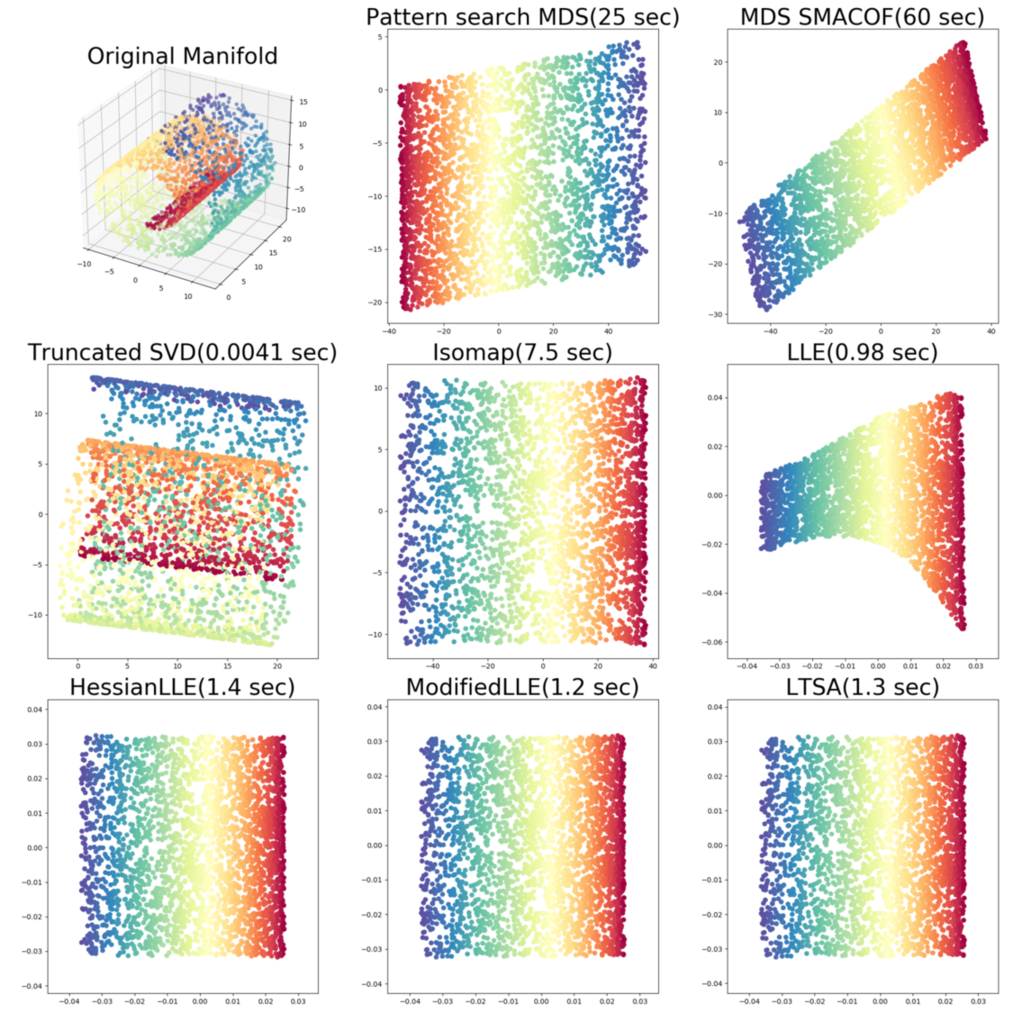}
  \caption{}
  \label{f:swissroll}
\end{subfigure}
\hfill
\begin{subfigure}[b]{.48\textwidth}
  \centering
  \includegraphics[width=\linewidth]{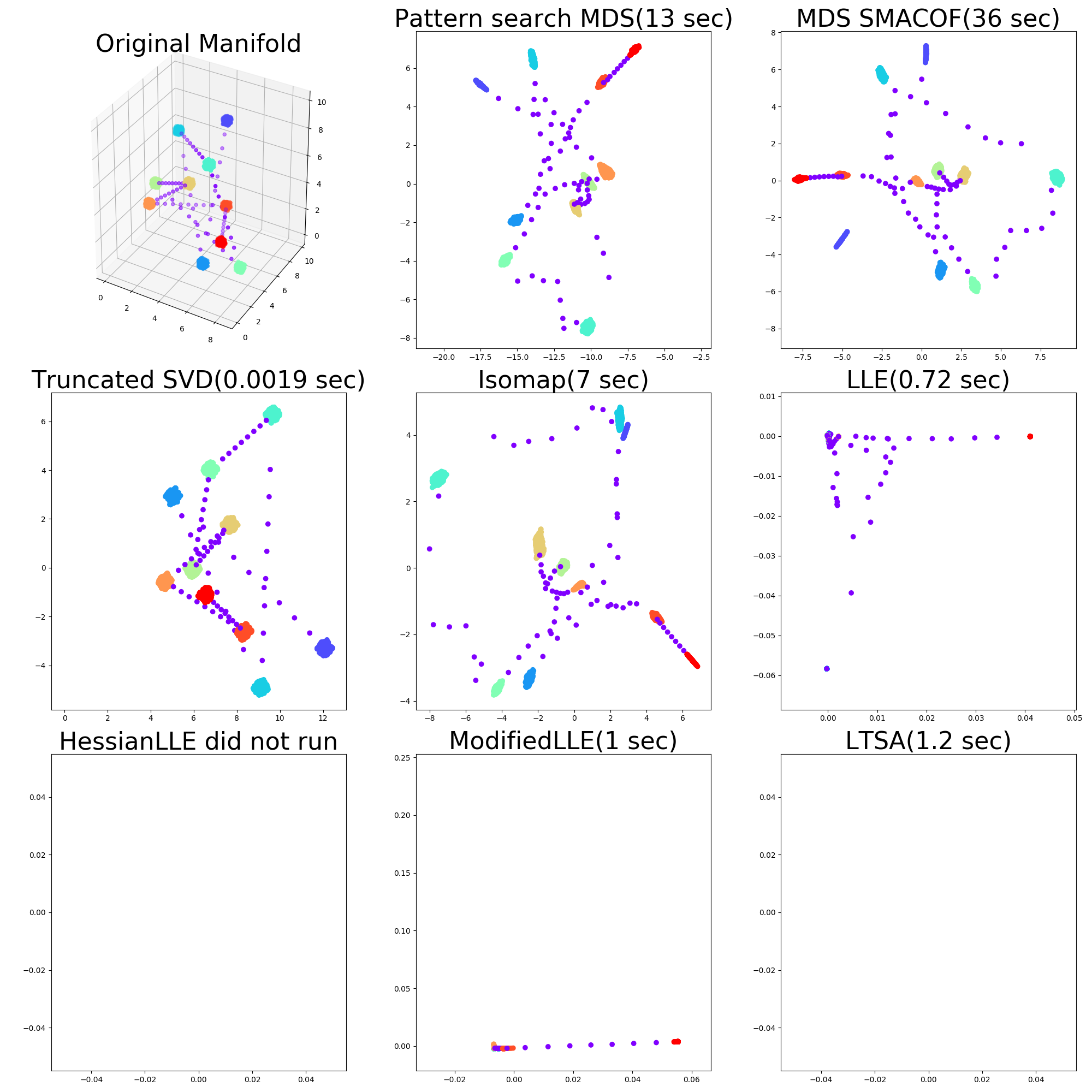}
  \caption{}
  \label{f:clusters}
\end{subfigure}

\begin{subfigure}[b]{.5\textwidth}
  \centering
  \includegraphics[width=\linewidth]{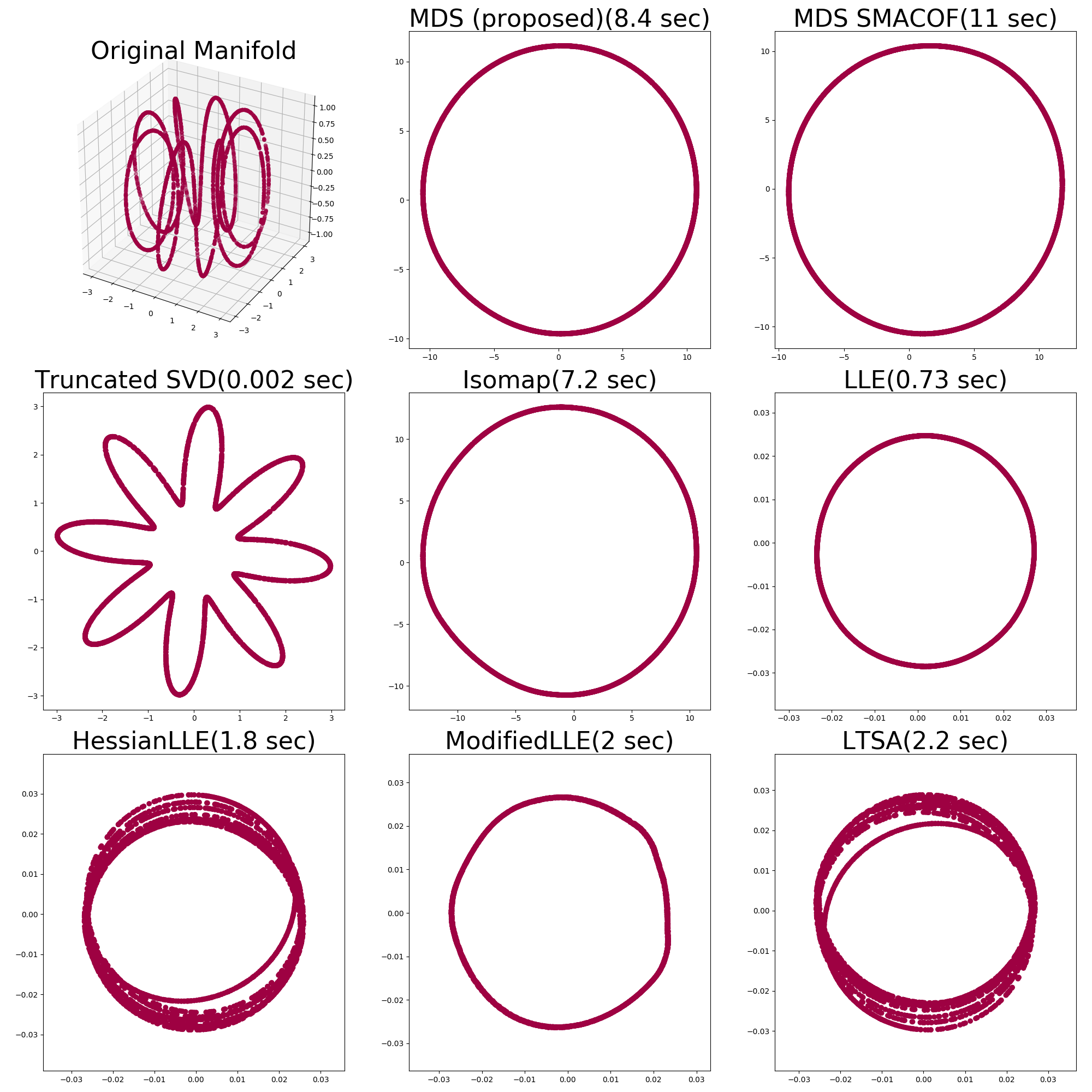}
  \caption{}
  \label{f:toroid-helix}
\end{subfigure}

\caption{
Comparison of pattern search MDS with other dimensionality reduction methods when converting: (\subref{f:swissroll}) $3D$ swissroll to $2D$ plane, (\subref{f:clusters}) $3D$ clusters to $2D$ clusters, and (\subref{f:toroid-helix}) $3D$ toroid helix to $2D$ circle
}
\end{figure*}

We present $3$ characteristic shapes selected from the ones we tested. The first shape we examine is the classical swissroll, where a $2D$ plane is ``rolled'' in  $3D$ space and the target is to extract the original $2D$ plane. Results are presented in Fig.~\ref{f:swissroll}. We observe that linear dimensionality reduction techniques like truncated SVD have trouble unrolling the swissroll. Also LLE introduces a lot of distortion to the constructed plane.

% \begin{figure}[h]
%   \centering
%   \includegraphics[width=\columnwidth]{swissroll}
% \caption{$3D$ swissroll to $2D$ plane comparison of proposed MDS with other dimensionality
%   reduction methods}
% \label{f:swissroll}
% \end{figure}

% \begin{figure}[h]
%   \centering
%   \includegraphics[width=\columnwidth]{twin_peaks}
% \caption{Twin Peaks comparison of proposed MDS with other dimensionality
%   reduction methods}
% \label{f:twin_peaks}
% \end{figure}

Next we examine how the algorithms handle sparse distance matrices. To this end, we generate a dataset of $3D$ non-overlapping clusters with a line connecting the centroids, where sparsity of the distance matrix follows because the vast majority of the points are very closely sampled inside the clusters. A good mapping should preserve the cluster structure in lower dimensions. In Fig.~\ref{f:clusters} we see that the truncated SVD and the MDS family of algorithms (proposed, SMACOF, Isomap) produce good results, while the LLE variants can't handle sparsity in distance matrices very well. In particular Hessian LLE and LTSA do not produce any output because of numerical instability\footnote{In Hessian LLE the matrices used for the null space computation become singular, while in LTSA the resulting point coordinates are infinite.} in the eigenvalue decomposition stages of these algorithms. Pattern search MDS does not rely on eigenvalue computation or equation system solvers and therefore it is numerically stable.

% \begin{figure}[h]
%   \centering
%   \includegraphics[width=\columnwidth]{clusters_3d}
% \caption{$3D$ cluster to $2D$ cluster comparison of proposed MDS with other dimensionality
%   reduction methods}
% \label{f:clusters}
% \end{figure}

Finally, we showcase how the algorithms perform with transitions from dense to sparse regions with a toroidal helix shape in Fig.~\ref{f:toroid-helix}. We can see that five methods, including pattern search MDS, unroll the shape into the expected $2D$ circle, while truncated SVD provides a daisy-like shape. Hessian LLE and LTSA collapse the helix into multiple overlapping circles.

% \begin{figure}[htb!]
%   \centering
%   \includegraphics[width=\columnwidth]{toroid-helix}
% \caption{$3D$ toroid helix to $2D$ circle comparison of proposed MDS with other dimensionality
%   reduction methods}
% \label{f:toroid-helix}
% \end{figure}

\subsection{Dimensionality reduction for semantic similarity}
\label{ss:sem-sim}

Construction of semantic network models consists of representing concepts as vectors in a, possibly high-dimensional, space $\mathbb{R}^n$. The relations between concepts are quantified as the distances, or inversely the cosine similarities, between semantic vectors. The semantic similarity task aims to evaluate the correlation of the similarities between concepts in a given semantic space against a set of ground truth similarity values provided by human annotators.

%Dimensionality reduction provides a way to reduce the size of the vectors resulting to significant gains in computation time and storage space. It is desired that a dimensionality reduction method preserves the original space structure, and specifically the dissimilarities between concepts.

%The MDS family of algorithms is a natural fit for this
% \thynote{No explanation -- delete or elaborate
% \geonote{ It is desired that a dimensionality reduction method preserves the original space structure, and specifically distances between concepts: By the definition of the problem MDS reconstructs a space where distances are preserved. We explain this about 100 times.}
% \thynote{Then do not restate the obvious
% \geonote{Touche :)}
% }
% }.  

We evaluate the performance of the dimensionality techniques investigated also in Section \ref{ss:geometry} for the semantic similarity task. We use the MEN \cite{Bruni:2014:MDS:2655713.2655714} and SimLex-999 \cite{hill2015simlex} semantic datasets as ground truth. Both datasets are provided in the form of lists of word pairs, where each pair is associated with a similarity score. This score was computed by averaging the similarities provided by human annotators. As the high-dimensional semantic word vectors, we use the $300$-dimensional GloVe vectors constructed by \cite{baziotis-pelekis-doulkeridis:2017:SemEval2} using a large Twitter corpus. We reduce the dimensionality of the vectors to the target dimension $L$  
% \potamnote{the choice of D 10,20,.. etc has to be explained and sample results should be provided for other Ds in the text}
and calculate the Spearman correlation coefficient between the human provided and the automatically computed similarity scores. Results are summarized in Table~\ref{t:semantic} for $L=10$. We observe that LLE yields the best results for MEN, while pattern search MDS performs best for SimLex-999. In addition, we observe that non-linear dimensionality reduction techniques can significantly improve the performance of the semantic vectors in some cases. %, specifically for SimLex-999.

\begin{table}[!htbp]
\centering
\renewcommand{\arraystretch}{1.4}
\begin{tabular}{|c|c|c|c|} \hline
 Dimensionality reduction & Dimensions & \textbf{MEN} & \textbf{SimLex-999}  \\ \hline\hline
 -                        &  $300$     & $0.635$ &  $0.177$         \\ \hline\hline
pattern search MDS        &  $10$      & $0.596$      &  $\textbf{0.242}$    \\ \hline
 MDS SMACOF               &  $10$      & $0.632$      &  $0.221$        \\ \hline
 Isomap                   &  $10$      & $0.625$      &  $0.132$             \\ \hline
 Truncated SVD            &  $10$      & $0.562$      &  $0.140$             \\ \hline
 LLE                      &  $10$      & $\textbf{0.657}$      &  $0.172$             \\ \hline
 Hessian LLE              &  $10$      & $0.157$      &  $0.004$             \\ \hline
 Modified LLE             &  $10$      & $0.643$      &  $0.158$             \\ \hline
 LTSA                     &  $10$      & $0.154$      &  $0.004$             \\ \hline 
 
\end{tabular}
\caption{Comparison of dimensionality reduction techniques for the semantic similarity task for MEN and SimLex-999 datasets.}
\label{t:semantic}
\end{table}

\subsection{Dimensionality reduction for k-NN classification}

The next set of experiments aims to compare the proposed algorithm to other dimensionality reduction methods for k-NN classification on a real dataset. We choose to use MNIST as a benchmark dataset which contains $70,000$ handwritten digit images.
We selected a random subset of $1000$ images and reduced the dimensionality from $784$ to $20$. Performance of the models is evaluated on 1-NN classification and using $10$-fold cross-validation. The evaluation metric is macro-averaged F1 score. Table~\ref{t:knn} summarizes the results. Observe that dimensionality reduction using pattern search MDS and Truncated SVD can improve classification performance over the original high-dimensional data. Pattern search MDS yields the best results overall. Hessian LLE, Modified LLE and LTSA did not run due to numerical instability.

\begin{table}[!htbp]
\centering
\renewcommand{\arraystretch}{1.4}
\begin{tabular}{|c|c|c|} \hline
 Method & Dimensions & \textbf{MNIST 1-NN F1 score}  \\ \hline\hline
 Original MNIST                        &  $784$     & $0.861$  \\ \hline\hline
 pattern search MDS          &  $20$      & $\textbf{0.878}$  \\ \hline
 MDS SMACOF               &  $20$      & $0.857$ \\ \hline
 Isomap                   &  $20$      & $0.829$ \\ \hline
 Truncated SVD            &  $20$      & $0.871$ \\ \hline
 LLE                      &  $20$      & $0.813$ \\ \hline
 Hessian LLE              &  $20$      & $-$ \\ \hline
 Modified LLE             &  $20$      & $-$ \\ \hline
 LTSA                     &  $20$      & $-$ \\ \hline 
 
\end{tabular}
\caption{Comparison of dimensionality reduction techniques for the MNIST dataset.}
\label{t:knn}
\end{table}

\subsection{Convergence characteristics}
\begin{figure}
\centering

\begin{subfigure}[b]{.48\textwidth}
  \centering
  \includegraphics[width=\linewidth]{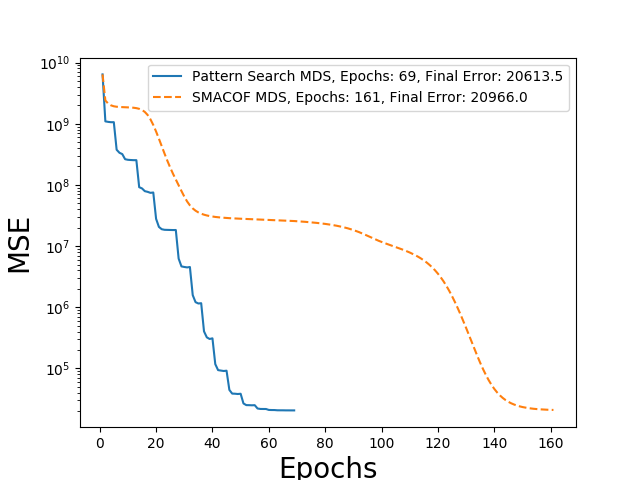}
  \caption{}
  \label{f:swissroll-convergence}
\end{subfigure}
%\hfill
\begin{subfigure}[b]{.48\textwidth}
  \centering
  \includegraphics[width=\linewidth]{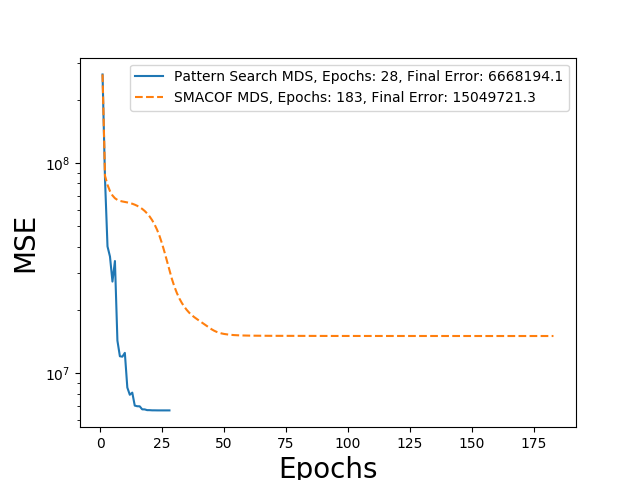}
  \caption{}
  \label{f:3d-clusters-convergence}
\end{subfigure}

\vfill
\begin{subfigure}[b]{.48\textwidth}
  \centering
  \includegraphics[width=\linewidth]{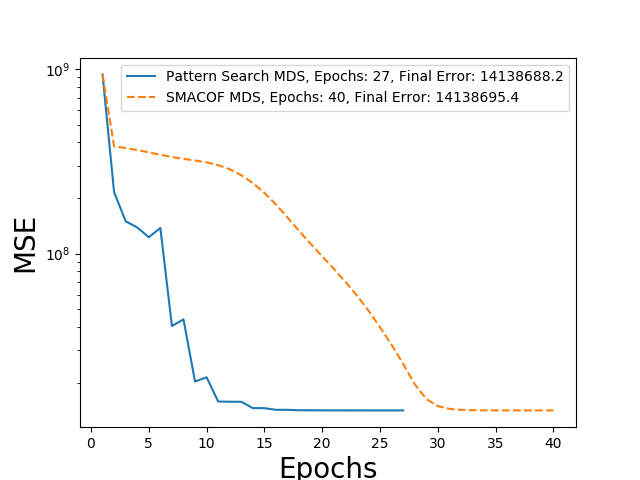}
  \caption{}
  \label{f:toroid-helix-convergence}
\end{subfigure}
%\hfill
\begin{subfigure}[b]{.48\textwidth}
  \centering
  \includegraphics[width=\linewidth]{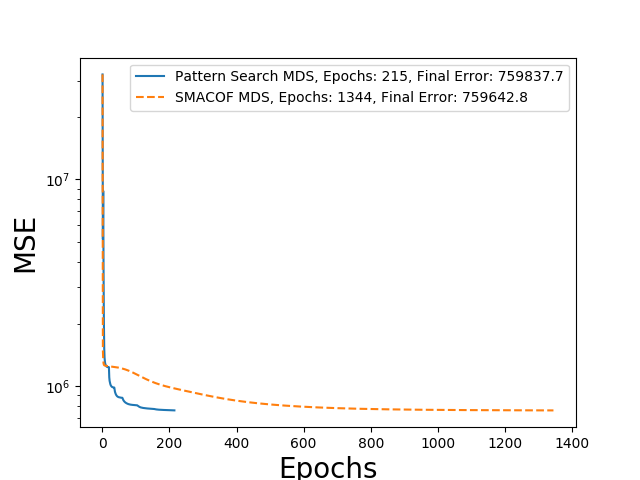}
  \caption{}
  \label{f:sem_sim_convergence}
\end{subfigure}

\caption{Convergence comparison of pattern search MDS and MDS SMACOF for (\subref{f:swissroll-convergence}) swissroll, (\subref{f:toroid-helix-convergence}) toroid helix, (\subref{f:3d-clusters-convergence}) 3d clusters and (\subref{f:sem_sim_convergence}) word vectors}
\end{figure}

Next we compare speed of convergence of pattern search MDS and MDS SMACOF, in terms of numbers of epochs. To this end we will consider the experiments of Sections \ref{ss:geometry} and \ref{ss:sem-sim} and present comparative convergence plots.  
We see the convergence plots for the cases of swissroll, $3D$ clusters, toroid helix in Fig.~\ref{f:swissroll-convergence},~\ref{f:3d-clusters-convergence} and~\ref{f:toroid-helix-convergence}, respectively.
The convergence plot for the word semantic similarity task is shown in Fig.~\ref{f:sem_sim_convergence}.
The plots are presented in y-axis logarithmic scale because the starting error is many orders of magnitude larger than the local minimum reached by the algorithms.

For all cases, we observe that pattern search MDS converges very quickly to a similar or better local optimum while MDS SMACOF hits regions where the convergence slows down and then recovers. These saw-like structure of the pattern search plots are due to the fact that we allow for ``bad moves'' as detailed in Section~ \ref{section: bad moves}.

% \geonote{ Needs rework. Be and every peak indicates a halving of the search radius. Another convergence characteristic of the proposed method is that we may allow the error to grow temporarily before we halve the radius which proved experimentally to help the algorithm converge faster and in better local optima. \thynote{Consider using instead of allow -- tolerance}}

\subsection{Robustness to noisy or missing data}

% \potamnote{How are the algorithms robust to missing data??? ... should be a new section or part of this one ...}

The final set of experiments aims to demonstrate the robustness of pattern search MDS when the input data are corrupted or noisy. To this end two cases of data corruption are considered: additive noise and missing data.

\subsubsection{Robustness to additive noise}

For this set of experiments, we inject Gaussian noise of variable standard deviation ($\sigma$) to the input data and use the dissimilarity matrix calculated on the noisy data as input to each one of the algorithms evaluated. %We compare the behavior of the proposed algorithm to the same dimensionality reduction techniques used in the previous sections 

% \thynote{Dimensionality reduction is the most frequent phrase could we reduce it or make an abbreviation DR?
% \geonote{If we are hard pressed with space I agree, else I don't think it helps readability}}.

\begin{figure}
\centering

\begin{subfigure}[b]{.49\textwidth}
  \centering
  \includegraphics[width=\linewidth]{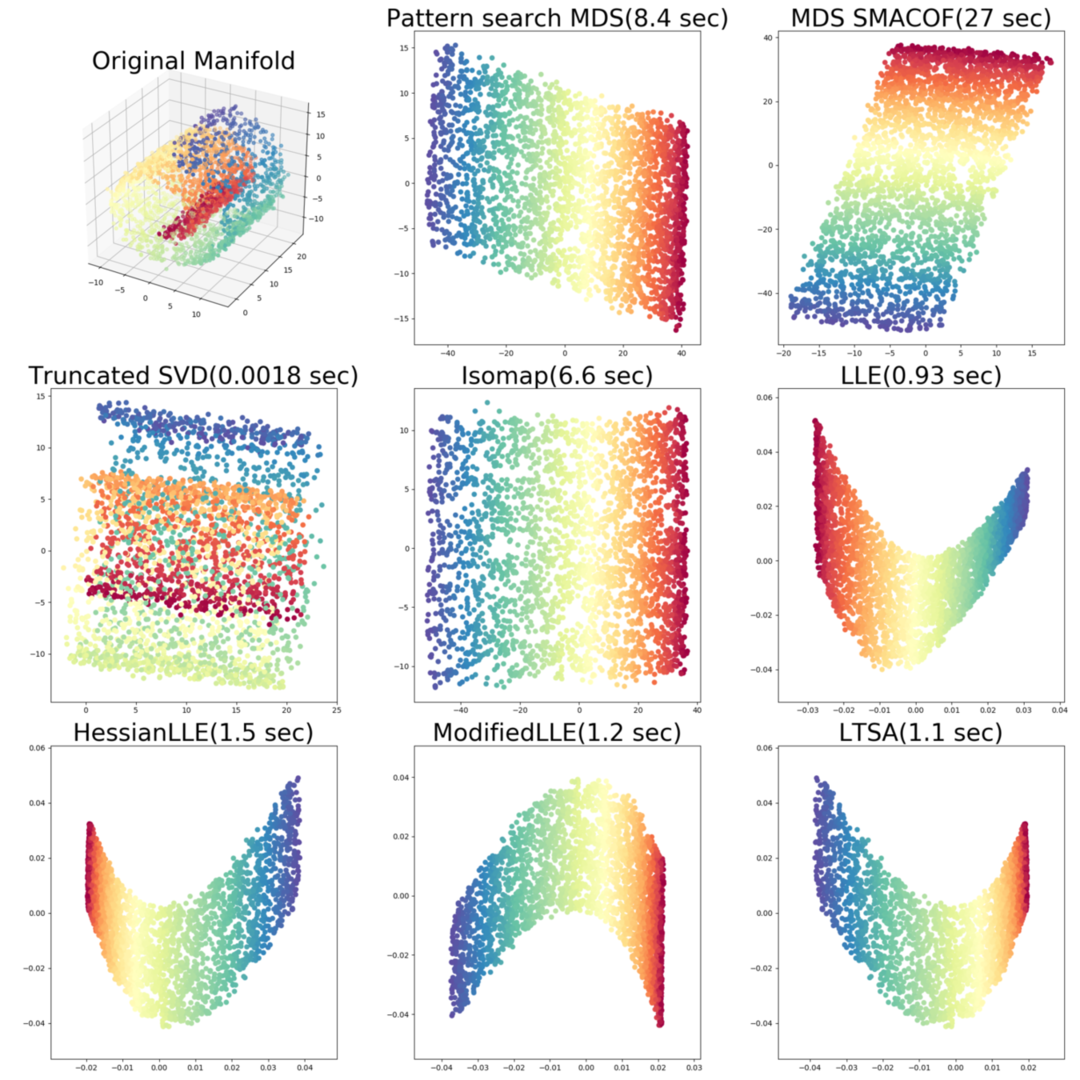}
\caption{}
\label{f:swissroll_noise}
\end{subfigure}
\hfill
\begin{subfigure}[b]{.49\textwidth}
  \centering
  \includegraphics[width=\linewidth]{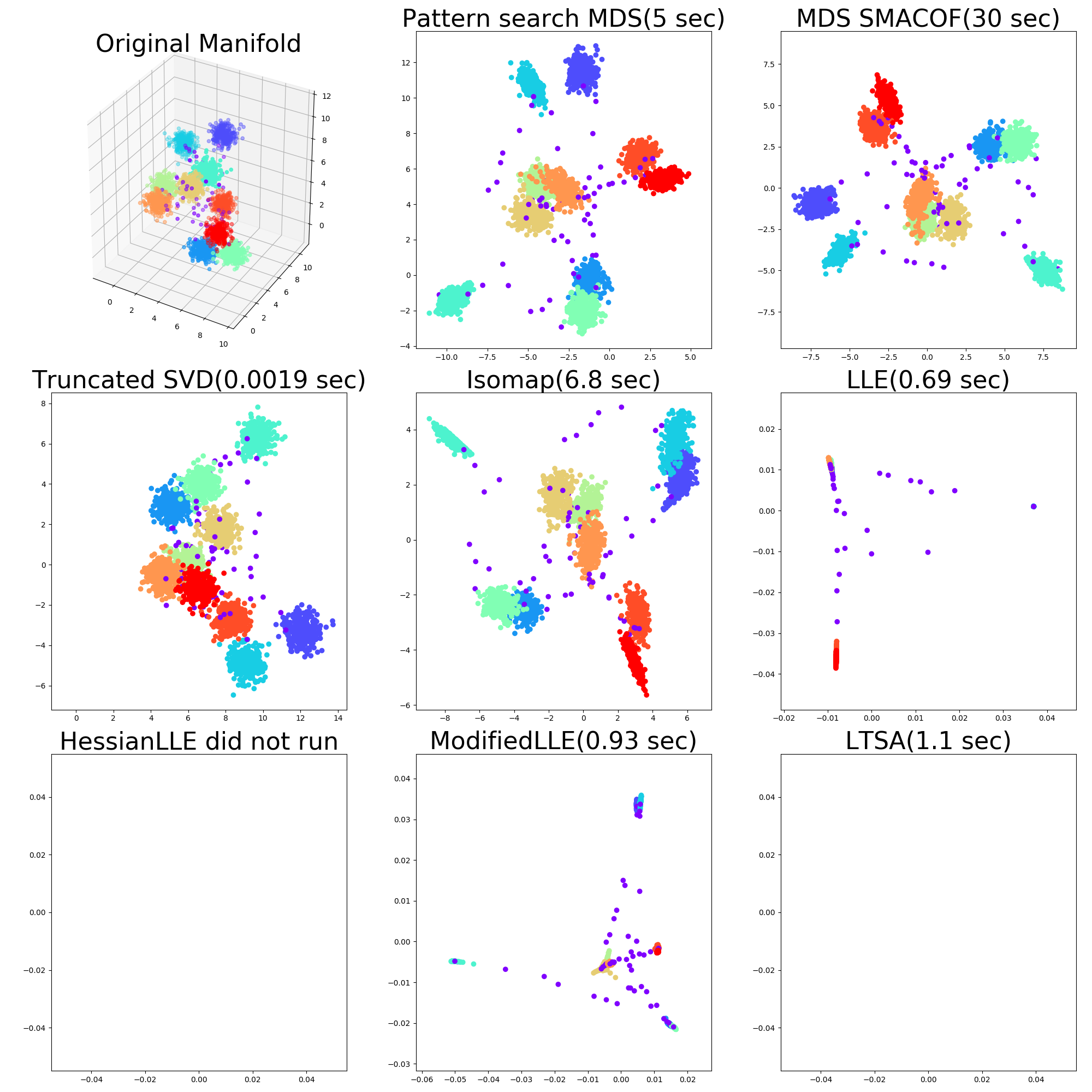}
\caption{}
\label{f:clusters_3d_noise}
\end{subfigure}

\begin{subfigure}[b]{.49\textwidth}
  \centering
  \includegraphics[width=\linewidth]{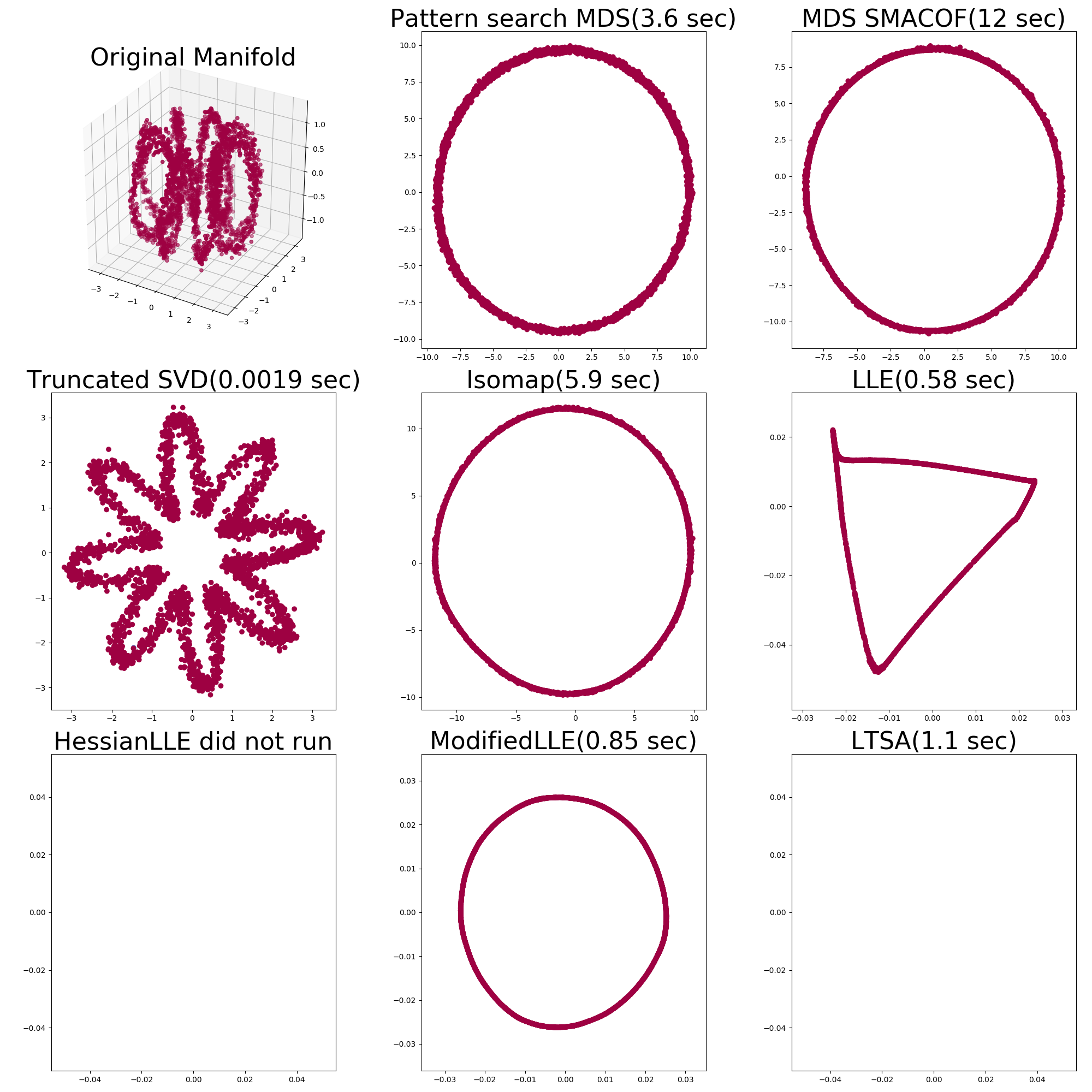}
\caption{}
\label{f:toroid_helix_noise}
\end{subfigure}

\caption{
Comparison of pattern search MDS with other dimensionality reduction methods when converting noisy (\subref{f:swissroll}) $3D$ swissroll to $2D$ plane ($\sigma=0.4$), (\subref{f:clusters}) $3D$ clusters to $2D$ clusters ($\sigma=0.3$) and (\subref{f:toroid-helix}) $3D$ toroid helix to $2D$ circle ($\sigma=0.07$)
}
\end{figure}

For the synthetic data of Section~\ref{ss:geometry}, we will follow a qualitative evaluation by showing the unrolled manifolds for high levels of noise. We perform dimensionality reduction for swissroll, toroid helix and $3D$ clusters for increasing noise levels. We report results for the highest possible noise deviation where one or more techniques still produce meaningful manifolds. Beyond these values of $\sigma$ the original manifolds become corrupted and the output of all methods is dominated by noise. 
Figs.~\ref{f:swissroll_noise},~\ref{f:clusters_3d_noise},~\ref{f:toroid_helix_noise}  show the results for noisy swissroll with $\sigma=0.3$, $3D$ clusters with $\sigma=0.4$ and toroid helix with $\sigma=0.07$ respectively. Overall, the pattern search MDS, followed by SMACOF MDS and Isomap are more robust to additive noise.

% \begin{figure}[htb!]
%   \centering
%   \includegraphics[width=\columnwidth]{swissroll_noise_4e-1}
% \caption{Comparison of dimensionality reduction techniques for noisy swissroll ($\sigma=0.4$)}
% \label{f:swissroll_noise}
% \end{figure}

% \begin{figure}[htb!]
%   \centering
%   \includegraphics[width=\columnwidth]{toroid_helix_noise_8e-2}
% \caption{Comparison of dimensionality reduction techniques for noisy toroid helix ($\sigma=0.08$)}
% \label{f:toroid_helix_noise}
% \end{figure}

% \begin{figure}[htb!]
%   \centering
%   \includegraphics[width=\columnwidth]{clusters_3d_3e-1}
% \caption{Comparison of dimensionality reduction techniques for noisy $3D$ clusters ($\sigma=0.3$)}
% \label{f:clusters_3d_noise}
% \end{figure}

For the semantic similarity task we injected different levels of Gaussian noise in the original word vectors and evaluated the correlation on MEN and Simlex-999. Results are presented in Table~\ref{t:semantic_noise}. We observe that the relative performance of the algorithms is maintained under noise injection, except for LLE which cannot handle high amounts of noise. LLE is achieving the best correlation values on MEN at $\sigma = 0.01$ and $\sigma = 0.1$, while pattern search MDS achieving the best performance on Simlex-999.

% \thynote{IN TABLE 3: I would bolt only the best results including the initial dataset for TABLE III as we show much more columns that we are first and in all the others no dim. reduction is first but this is minor}

\begin{table}
    \begin{adjustwidth}{-.5in}{-.5in}
        \begin{center}
\renewcommand{\arraystretch}{1}
\footnotesize
\begin{tabular}{|c|c|c|c|c|c|c|c|} \hline
 Method & Dimensions & \multicolumn{3}{c|}{\textbf{MEN}} & \multicolumn{3}{c|}{\textbf{SimLex-999}}  \\ \hline\hline 
 && $\sigma=0.01$ & $\sigma=0.1$ & $\sigma=0.5$ & $\sigma=0.01$ & $\sigma=0.1$ & $\sigma=0.5$ \\ \hline\hline
 Original GloVe                        &  $300$ & $0.635$ & $0.619$ & $0.431$ & $0.178$ & $0.169$ & $0.077$   \\ \hline
 pattern search MDS       &  $10$  & $0.593$ & $0.597$ & $0.462$ & $\textbf{0.249}$  & $\textbf{0.315}$ & $\textbf{0.204}$   \\ \hline
 MDS SMACOF               &  $10$  & $0.633$ & $0.620$ & $0.462$ & $0.229$  & $0.222$ & $0.123$        \\ \hline
 Isomap                   &  $10$  & $0.622$ & $0.613$ & $\textbf{0.497}$ & $0.134$ & $0.124$ & $0.079$            \\ \hline
 Truncated SVD            &  $10$  & $0.562$ & $0.551$ & $0.380$ & $0.140$ & $0.136$ & $0.039$            \\ \hline
 LLE                      &  $10$  & $\textbf{0.659}$ & $\textbf{0.649}$ & $0.369$ & $0.175$ & $0.166$ & $0.052$          \\ \hline
 Hessian LLE              &  $10$  & $0.156$ & $0.144$ & $0.023$ & $0.005$ & $0.04$ & $0.018$           \\ \hline
 Modified LLE             &  $10$  & $0.635$ & $0.633$ & $0.489$ & $0.158$ & $0.162$ & $0.096$          \\ \hline
 LTSA                     &  $10$  & $0.155$ & $0.141$ & $0.020$ & $0.06$ & $0.04$ & $0.002$            \\ \hline 
 
\end{tabular}
\caption{Comparison of dimensionality reduction techniques with noisy word vectors on the semantic similarity task for MEN and SimLex-999 datasets.}
\label{t:semantic_noise}
        \end{center}
    \end{adjustwidth}
\end{table}

\subsubsection{Robustness to missing data}
\label{ss:missing}
\begin{figure}
\centering

\begin{subfigure}[b]{.49\textwidth}
  \centering
  \includegraphics[width=\linewidth]{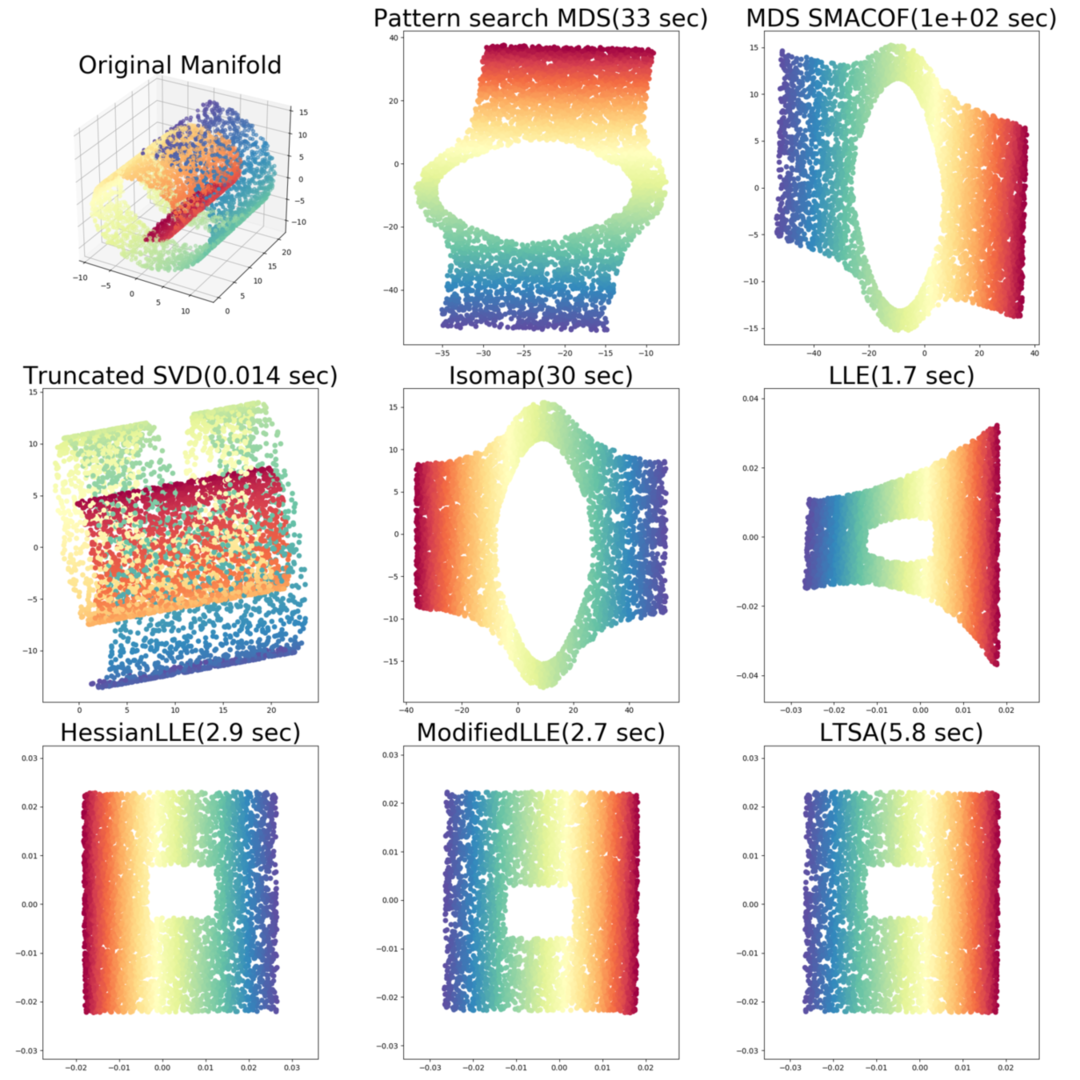}
\caption{}
\label{f:swisshole}
\end{subfigure}
\hfill
\begin{subfigure}[b]{.49\textwidth}
  \centering
  \includegraphics[width=\linewidth]{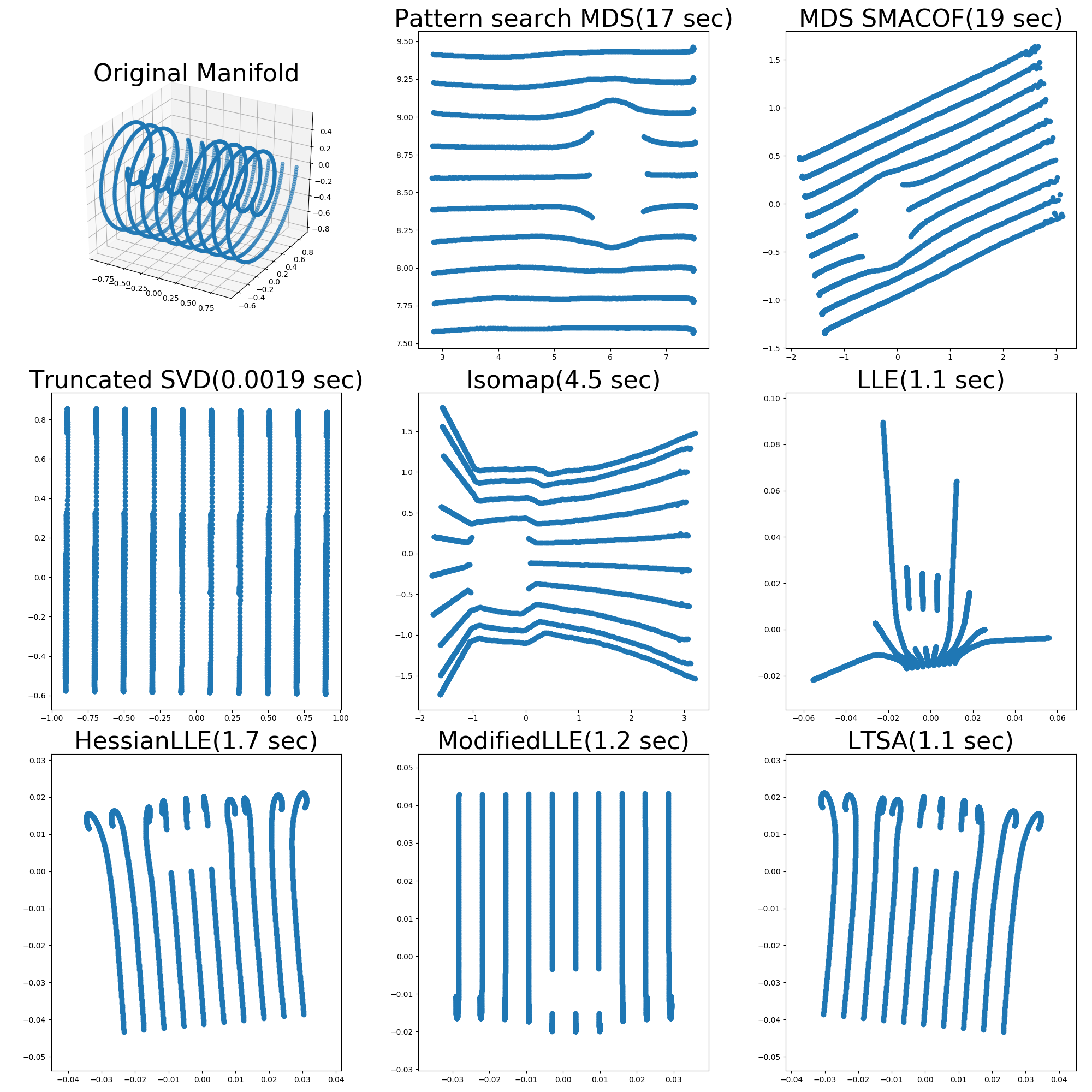}
\caption{}
\label{f:spiral_hole}
\end{subfigure}

\caption{
Comparison of pattern search MDS with other dimensionality reduction methods for (\subref{f:swisshole}) dense and (\subref{f:spiral_hole}) sparse swissroll with hole. Target is a plane with a rectangular hole.}
\label{f:holes}
\end{figure}

For the final set of experiments we consider the case of missing data. For this two new synthetic datasets where constructed, namely a dense and a sparse swissroll with a hole as shown in Fig.~\ref{f:holes}. In Fig~\ref{f:swisshole}, we show the performance of the various algorithms applied to a dense swissroll with a hole in the middle. As we can see only Hessian LLE, modified LLE and LTSA are able to reconstruct the shape correctly, while MDS algorithms result in distortion around the hole. This is due to the non-convexity we introduced to the space when adding the hole. This distortion can still be observed (to a lesser degree) in the sparse variation shown in Fig.~\ref{f:spiral_hole}. For the sparse data case, we observe that LLE methods result in distortion around the edges.

These preliminary experiments indicate that LLE variations can handle better non-convexities in input data, while MDS variations can handle sparse data better. This is because LLE methods are based on inferring and combining local data geometry, while MDS methods are inferring global geometry. 

\section{CONCLUSIONS\label{section:CONCLUSIONS}}
We propose pattern search MDS, a novel algorithm for nonlinear dimensionality reduction, inspired by gradient-free optimization methods. Pattern search MDS is formulated as an instance of the wider  family of GPS methods, thus providing theoretical guarantees of convergence up to a fixed point. Additional optimizations further improve the performance of our algorithm in terms of computational efficiency, robustness and solution quality. The qualitative evaluation against other popular dimensionality reduction techniques for both clean and noisy manifold geometry shapes indicates that pattern search MDS can accurately infer the intrinsic geometry of manifolds embedded in high-dimensional spaces. Furthermore, the comparison of convergence characteristics against SMACOF MDS show that pattern search MDS converges in fewer epochs to similar or better solutions. Experiments on real data yield comparable to state-of-the-art results both for a lexical semantic similarity task  and on MNIST for KNN classification. Open-source implementations of pattern search MDS and the data generation process are provided to facilitate the reproducibility of our results.

%We propose a novel algorithm for NLDR inspired by gradient-free optimization. The reduction of our algorithm as an instance of the wider family of GPS methods enable us to prove the convergence up to a fixed-point. Additional optimizations further improve the performance of our algorithm in terms of computational efficiency, robustness and accuracy. Experiments clean and noisy environments yield state-of-the-art results in both synthetic and real high dimensional data.

% \potamnote{missing :)
% \thynote{Added a small one
% \geonote{Added a bit larger one :)}
% }}

\section{FUTURE WORK\label{section:FUTURE WORK}}

Future work will focus on improving runtime performance and scalability of pattern search MDS.
Specifically, an approach for decreasing per epoch computational complexity is to narrow the search space of possible moves as the geometry of the embedding space becomes more apparent by biasing the moves towards the principal component vectors of the neighborhood of the point that is being moved. This can be viewed as a combination of pattern search and gradient descent, where the search space of moves is wide at the beginning and then gets increasingly biased towards the direction of the gradient. Our algorithm can scale to large numbers of points by utilizing Landmark points \cite{silva2004-landmark} or fast approximations to MDS \cite{Yang06afast}. These approaches aim to alleviate the computational and memory cost of computing the full distance matrix, by approximating the data geometry using smaller submatrices. Moreover, stochastic approximations like stochastic SMACOF \cite{rajawat2017stochastic} can be adapted to pattern search MDS. 

We also plan to provide more in-depth theoretical insights and ways to enable pattern search MDS to capture complex geometrical properties of input data. We aim to perform a detailed analysis on how heuristics and especially allowing for ``bad moves'' affect the performance of pattern search MDS. Furthermore, in Sections \ref{ss:geometry} and \ref{ss:missing} we showcased that MDS can better handle sparse data and LLE can better handle non-convexity and missing data. This makes sense, as MDS takes into account the global geometry of the embedding space, while LLE focuses on the geometry of local neighborhoods. We plan to combine the cost functions of these approaches to infer both global and local geometry of the low dimensional data manifold. Another way to increase the expressiveness of the algorithm is to investigate a wider variety of distance metrics, and specifically non-symmetrical distance ``metrics'', motivated by cognitive sciences \cite{pothos2011quantum}.

\addtolength{\textheight}{-12cm}   % This command serves to balance the column lengths

%%%%%%%%%%%%%%%%%%%%%%%%%%%%%%%%%%%%%%%%%%%%%%%%%%%%%%%%%%%%%%%%%%%%%%%%%%%%%%%%

%%%%%%%%%%%%%%%%%%%%%%%%%%%%%%%%%%%%%%%%%%%%%%%%%%%%%%%%%%%%%%%%%%%%%%%%%%%%%%%%

%%%%%%%%%%%%%%%%%%%%%%%%%%%%%%%%%%%%%%%%%%%%%%%%%%%%%%%%%%%%%%%%%%%%%%%%%%%%%%%%
%\section*{APPENDIX\label{section:APPENDIX}}

% \potamnote{Funding ... and some ppl that wrote code ... georgia kostantinos ...
% \thynote{I do not know how to properly say thanks to someone who wrote code we did not use :P}}

%%%%%%%%%%%%%%%%%%%%%%%%%%%%%%%%%%%%%%%%%%%%%%%%%%%%%%%%%%%%%%%%%%%%%%%%%%%%%%%

\section*{ACKNOWLEDGMENTS\label{section:ACKNOWLEDGMENTS}}
This work has been partially supported by the EU-IST H2020 BabyRobot project under grant \# 687831. Special thanks to Nikolaos Ellinas for his productive feedback and Georgia Athanassopoulou and Konstantinos Mitropoulos for contributing to initial versions of the MDS code.

\addtolength{\textheight}{12cm}

\printbibliography

\end{document}